\documentclass[twocolumn, switch, a4paper]{article} 

\usepackage{preprint}

\usepackage{amsmath, amsthm, amssymb, amsfonts}

\usepackage{mathptmx}

\usepackage{verse}
\usepackage{metrix} 
\usepackage[LGR,T1]{fontenc}  
\usepackage[utf8]{inputenc}	

\usepackage{xcolor}		
\usepackage[colorlinks = false,
            linkcolor = purple,
            urlcolor  = blue,
            citecolor = cyan,
            anchorcolor = black]{hyperref}	
\usepackage{booktabs,tabularx,graphicx} 		
\usepackage{nicefrac}		
\usepackage{microtype}		
\usepackage{lineno}		
\usepackage{float}			
\usepackage{adjustbox}
\usepackage[hang,flushmargin]{footmisc} 
\usepackage{placeins} 

\usepackage{apacite} 
\AtBeginDocument{} 
\usepackage{etoolbox}

\AtBeginEnvironment{APACrefURL}{\renewcommand{\url}[1]{}} 
\renewcommand{\doiprefix}{doi:~\kern-1pt} 

\usepackage{lipsum}		

\usepackage{newfloat}
\usepackage{subfigure}
\DeclareFloatingEnvironment[name={Supplementary Figure}]{suppfigure}
\usepackage{sidecap}
\sidecaptionvpos{figure}{l}
\usepackage{caption}
\usepackage{titlesec}
\titlespacing*{\section}{0pt}{0.4\baselineskip}{0.35\baselineskip}
\titlespacing*{\subsection}{0pt}{0.4\baselineskip}{0.25\baselineskip}
\titlespacing*{\subsubsection}{0pt}{0.4\baselineskip}{0.25\baselineskip}

\usepackage{changepage}

\usepackage{tikz,xcolor,hyperref}

\definecolor{lime}{HTML}{A6CE39}
\DeclareRobustCommand{\orcidicon}{
  \begin{tikzpicture}
  \draw[lime, fill=lime] (0,0) 
  circle [radius=0.16] 
  node[white] {{\fontfamily{qag}\selectfont \tiny ID}};
  \draw[white, fill=white] (-0.0625,0.095) 
  circle [radius=0.007];
  \end{tikzpicture}
  \hspace{-2mm}
}
\foreach \x in {A, ..., Z}{\expandafter\xdef\csname orcid\x\endcsname{\noexpand\href{https://orcid.org/\csname orcidauthor\x\endcsname}
      {\noexpand\orcidicon}}
}

\usepackage{setspace} 
\title{Some Stylometric Remarks on Ovid's \emph{Heroides} and the \emph{Epistula Sapphus}}

\usepackage{xwatermark}
\newwatermark[firstpage,color=gray!90,angle=0,scale=0.28, xpos=0in,ypos=-5in]{correspondence: \texttt{benjamin.nagy@ijp.pan.pl}}

\usepackage{authblk}

\author{Ben Nagy\orcidA{}}
\affil{
Institute of Polish Language, Polish Academy of Sciences (IJP PAN)
}

\begin{document}
\twocolumn[ 
  \begin{@twocolumnfalse} 
  
\maketitle

\begin{abstract}
This article aims to contribute to two well-worn areas of debate in classical
Latin philology, relating to Ovid's \emph{Heroides}. The first is the question
of the authenticity (and, to a lesser extent the correct position) of the
letter placed fifteenth by almost every editor---the so-called \emph{Epistula
Sapphus} (henceforth \emph{ES}). The secondary question, although perhaps now
less fervently debated, is the authenticity of the `Double \emph{Heroides}',
placed by those who accept them as letters 16--21. I employ a variety of
methods drawn from the domain of computational stylometry to consider the
poetics and the lexico-grammatical features of these elegiac poems in the
broader context of a corpus of `shorter' (from 20 to 546 lines) elegiac works
from five authors (266 poems in all) comprising more or less all of the
non-fragmentary classical corpus. Based on a variety of techniques, every
measure gives clear indication that the poetic style of the \emph{Heroides} is
Ovidian, but distinctive; they can be accurately isolated from Ovid more
broadly. The Single and Double \emph{Heroides} split into two clear groups,
with the \emph{ES} grouped consistently with the single letters. Furthermore,
by comparing the style of the letters with the `early' (although there are
complications in this label) works of the \emph{Amores} and the late works of the
\emph{Ex Ponto}, the evidence supports sequential composition---meaning that
the \emph{ES} is correctly placed---and, further, supports the growing
consensus that the double letters were composed significantly later, in
exile.\footnotemark

\end{abstract}  %
\vspace{0.35cm}

\end{@twocolumnfalse} 
] 


    \footnotetext{Parts of this article were given as a paper in the seminar
    series of the department of Classics, Archaeology and Ancient History at
    the University of Adelaide in October 2021, and I am grateful to the
    participants for their useful comments and advice. Further thanks are due
    to Prof. Peter Davis for a variety of suggestions that have undoubtedly
    improved the article.}

\section{Introduction}

In form, Ovid's \emph{Heroides} are a collection of twenty-one elegies of
medium length.%
    \footnote{ About 190 lines, on average, significantly longer than the
    average length of the \emph{Amores} (50), \emph{Tristia} (71) or \emph{Ex
    Ponto} (69), but not `book length' like the \emph{Fasti} or \emph{Ars
    Amatoria}.}
They are written in epistolary style; the poet ventriloquises fifteen women
from mythology, before finishing with three pairs comprising a letter from a
`hero' and then a response (the so-called `Double' \emph{Heroides}). The sole
exception to this pattern is a letter from Sappho (who is not mythological) to
Phaon (who probably is). By those who doubt its authenticity, this is known as
the \emph{Epistula Sapphus}, but it appears in most modern editions as
\emph{Heroides} 15. As works of literature, the \emph{Heroides} are
exceptionally important for their presentation of the events of typically
male-centred genres, primarily epic and tragedy, from a woman's perspective.
While still classified as love elegy (they are about love and are written in
elegiac couplets), they reverse the male gaze and allow readers to assume a
female point of view. In terms of characterisation, Ovid in the
\emph{Heroides} dons distinct foreign personalities, yielding an obvious and
useful point of comparison to the \emph{praeceptor amoris} of the \emph{Ars
Amatoria}.%
    \footnote{Cf Catullus 16: \emph{nam castum esse decet pium poetam | ipsum,
    versiculos nihil necesse est}, `an upstanding poet should be virtuous
    themselves, but their verses need not be'.}
From a philological perspective, which is the one taken in this article, the
poems raise a number of questions, the most important ones being the
composition dates (are they early or late?) and the authenticity of several of
them---for not only the \emph{ES}, but all of the Double \emph{Heroides} have
been athetised by various scholars.

While these questions have been discussed at length by many excellent literary
scholars (a summary appears in the following section), a modern stylometric
treatment is overdue, and that is what this article will attempt to provide.
Following a reprise of the key issues, and the standard arguments for and
against, I present the results of a computational stylometric analysis that
examines two broad domains: lexico-grammatical style (comprising word choice,
phonetic preferences, and grammatical style) and `poetic' style (which ignores
lexis completely, considering only issues of prosody, rhyme and assonance).
Based on a variety of techniques, every measure gives clear indication that
the poetic style of the \emph{Heroides} is Ovidian, but distinctive; they can
be accurately isolated from Ovid more broadly. The single and double
\emph{Heroides} split into two clear groups, with the \emph{ES} grouped
consistently with the single letters. Furthermore, by comparing the style of
the letters with the `early' (although there are complications in this label)
works of the \emph{Amores} and the late works of the \emph{Ex Ponto}, the
evidence supports sequential composition---meaning that the \emph{ES} is
correctly placed---and, further, supports the contention that the double
letters were composed significantly later, in exile.

\section{The Bones of the Existing Debate}
\label{sec:bones}

\subsection{The Authorship of the \emph{ES}}

As space is limited, here I present only the bones of the existing debate,
aiming mostly to establish some context and offer a few opinions on certain
points. For a complete history that is fairly up to date I can recommend no
better source than \citeA[97--122]{thorsen_ovearly}, which includes a
conveniently detailed literary response to \citeA{tarrant_81}, whose article
is the linchpin of the modern case against the \emph{ES}.

The first thing that must be said is that the transmission of the \emph{ES} is
troubled, even in the context of the relatively poor transmission of the
\emph{Heroides} in general (Ovid more broadly survives in generally excellent
condition). Of the oldest manuscripts, the full text of the \emph{ES} appears
only in the thirteenth century \emph{Francofurtanus} (MS Barth. 110), although
the letter must have been widely known in the middle ages, since excerpts
survive in at least six Medieval \emph{florilegia}. However after 1420 we have
a profusion of manuscripts descended from a lost second source. For a detailed
conspectus and discussion, see \citeA[287ff.]{dorrie_71} (in Latin) or prefer
\citeA{tarrant_trans_ep} for concision.

The \emph{ES} was, in 1629, placed fifteenth in the collection by Daniel
Heinsius, following the notes of Scaliger.%
    \footnote{In \textbf{F} it appears before the rest, but the letters in
    that MS are jumbled, see \citeA[288]{dorrie_71}. For the full story,
    \citeA[21--2]{thorsen_ovearly}.}
Scaliger believed it genuine (but doubted the Double \emph{Heroides}), as
did Heinsius. It was first doubted in 1816 by Francke for mistaken reasons,
then by Schneidewin who proposed, but later retracted, the idea that it was
a Renaissance forgery. Really, though, the seminal doubts were planted by
Lachmann, a towering Latinist, in 1848. His arguments convinced Palmer who,
in his 1874 edition, omitted the poem since it was ``condemned by Lachmann,
and \emph{every scholar possessed of common sense}'' (my emphasis). Years of
heated debate followed%
    \footnote{\citeA[102 n. 28]{thorsen_ovearly} finds at least nineteen
    articles in the intervening two decades!}
and by 1898 Palmer had reassessed the arguments, so that when concluding the
second edition he instructed his successor to ``defend as far as possible
[its] Ovidian authorship''. Although I have glossed over the details,
suffice it to say that the points of debate up to this point were both minor
and few, and could easily be put down to problems with transmission.

The time has now come to address the elephant in the room. Those disputing the
authenticity of the \emph{ES} need to deal with one very good reason to
believe it to be genuine---the fact that \emph{Ovid himself tells us he wrote
it}. The relevant passage%
    \footnote{In fact there are two references to Sappho in \emph{Am}. 2.18; I
    do not focus on the second because it might be thought to refer to one of
    the works of Sabinus, who was evidently inspired by Ovid's letters to
    compose some male replies.}
is from \emph{Amores} 2.18:
\begin{adjustwidth}{0.07\linewidth}{0.07\linewidth}
  \footnotesize
  \smallskip
  \begin{onehalfspace}
    aut, quod Penelopes verbis reddatur Ulixi,\\
    \phantom{xx} scribimus[\dots\\
    \dots]\\
    quodque tenens strictum Dido miserabilis ensem\\
    \phantom{xx} dicat et \dag Aoniae Lesbis amata lyrae\dag.%
    \footnote{The reading here is intractable, but no proposed version does
    away with the fact that it is a reference to Sappho.}
    \begin{flushright}
      Ov. \emph{Am}. 2.18.21--6
    \end{flushright}
  \end{onehalfspace}
  \phantom{xx}\\
  I write what Penelope's words conveyed to Ulysses [\dots several
  references to other \emph{Heroides}\dots] and what poor Dido, holding
  a drawn sword, might say, or the lover from Lesvos with the Aeonian lyre.
\end{adjustwidth}
Mostly this problem is handled in one of two ways---either by claiming that
Ovid did write a letter from Sappho, but that it was lost, so the present
\emph{ES} is a replacement by an interpolator,%
    \footnote{Thus eg \citeA[398]{mckeown_98} while acknowledging that it
    ``puts an agonising strain on credibility''.}
or alternatively by arguing that the \emph{ES} was never genuine and the lines
in \emph{Amores} 2.18 have been tampered with to authenticate it.
\citeA{tarrant_81} chooses the latter path, but Courtney's
\citeyear[163]{courtney_97} rebuttal seems unassailable (emphasis mine):

\begin{adjustwidth}{0.07\linewidth}{0.07\linewidth}
\footnotesize%
 Suppose we want to declare the Letter of Sappho spurious; we then run up
 against the difficulty that Ovid himself twice refers to it in \emph{Am.} 2.18
 \dots We then have to presume that the author of the Letter of Sappho
 validated his forgery by rewriting two lines of \emph{Am.} 2.18 to introduce
 mention of it; \emph{how did he then impose his will on the whole textual
 tradition?}
\end{adjustwidth}

As for the rest of Tarrant's arguments, I doubt that I could add much to
Thorsen's \citeyear[105 ff.]{thorsen_ovearly} point-by-point response and in
any case that is not the aim of this article. Tarrant, as is common in
Classical scholarship, focuses his attention on \emph{rare events}---a
metrical oddity here, an unusual collocation of words there, a turn of phrase
that is not attested until Neronian times. In stylometric (and statistical)
terms---and this will be discussed in more depth in \S \ref{sec:
methodology}---I am more concerned with analysis that considers \emph{common}
events, like the prosodic features of every line, or the relative frequencies
of thousands of different words. In fact \citeA[272]{tarrant_trans_ep} himself
seems to acknowledge that further analysis was needed: ``\dots a careful
stylistic analysis of the collection has not yet been undertaken and the
question therefore remains open'' (although here he refers more to the
authorship of the Double \emph{Heroides}, which are discussed next). I do not
at all dismiss the approach taken in Tarrant's article (although I might
debate the specifics); the point I am making is that computational assistance
opens new and different avenues of investigation by allowing us to
simultaneously consider thousands of stylistic markers in a way that is simply
impossible for humans.

\subsection{The Double \emph{Heroides}}

Despite his claim to the contrary, Tarrant, in questioning the authenticity of
the \emph{ES}, is influenced to some extent by his perception of its
quality---``[i]t is my private opinion that the ES is a tedious production
containing hardly a moment of wit, elegance, or truth to nature, and that its
ascription to Ovid ought never to have been taken seriously''
\cite[134--5]{tarrant_81}. Indeed readers have often (and perhaps
understandably) thought of the Single \emph{Heroides} that ``[i]t is difficult
to rescue them, especially if read sequentially from the charge of monotony''
\cite[1]{kenney_96}.%
    \footnote{\citeA[106]{wilkinson_74} famously compared them to a glut of
    plum pudding: ``The first slice is appetising enough, but each further
    slice becomes colder and less digestible until the only incentive for
    going in is the prospect of coming across an occasional ring or
    sixpence.''}
For the Double \emph{Heroides}, however, the case is quite different. Few
modern critics will deny that their style is quintessentially Ovidian%
    \footnote{\citeA[20]{kenney_96} cites Rand (1925) with approval: ``If [the
    double letters] are not from Ovid's pen, an \emph{ignotus} has beaten him
    at his own game.''}
nor that they are, in short, very good.%
    \footnote{Eg \citeA[330 n. 1]{reeve_heroides}: ``Lachmann's observations
    seem to me much too weak to establish the existence of a second poet as
    talented as Ovid, or more talented, it might be thought, than the Ovid of
    \emph{Epp}. 1--15''; even \citeA[157]{courtney_97} (contra) is forced to
    concur: ``[l]et me make no bones about assenting to the general
    verdict''.}
By varying the format, the poet profited from new opportunities for drama,
development and characterisation.

So what, then, is the sticking point? There are a few questions of diction and
style, and some problems with the transmission of large sections of letters 16
and 21,%
    \footnote{See \citeA[esp. 142--5]{heyworth_16} with references for an
    efficient and up to date précis.}
but the most urgent concern, which may surprise non-specialists, relates to
word length. In Augustan elegy (which is composed in dactylic couplets, one
hexameter then one pentameter) it is conventional to end the pentameter with a
disyllabic word. Ovid in particular is absolutely punctilious about this in
his early verse (the other poets less so), but he varies his style in the
exile poetry, permitting more and greater exceptions over time.%
    \footnote{\citeA[160--1]{courtney_97} summarises Ovid's development.
    Interestingly, although not strictly relevant, Propertius went in the
    opposite direction, permitting all sorts of endings in his early works but
    becoming more disyllabic over time.}
The Double \emph{Heroides}, once assumed (if genuine) to be composed
alongside the single letters, contain three polysyllabic endings. The
modern debate has run something like this:

\begin{adjustwidth}{0.07\linewidth}{0.07\linewidth}
\footnotesize%
[W]hen we take into account the development towards polysyllabic
pentameter-endings indicated above\dots it will follow that these pairs of
letters cannot be genuine, for it is inconceivable, and conceived by no-one,
that Ovid could have written poetry of this type at Tomis.
\begin{flushright}
\cite[64]{courtney_65}
\end{flushright}
\end{adjustwidth}

\begin{adjustwidth}{0.07\linewidth}{0.07\linewidth}
\footnotesize%
Courtney\dots points out that parallels can be found only in the poems from
exile; but why should Ovid not have composed \emph{Epp}. 16--21 in exile?
\begin{flushright}
\cite[330 n. 1]{reeve_heroides}
\end{flushright}
\end{adjustwidth}

\begin{adjustwidth}{0.07\linewidth}{0.07\linewidth}
\footnotesize%
If we assume that they are late\dots the anomalies stop worrying us. The only
price to pay would be to assume that Ovid is not very honest when he claims to
have given up writing light elegiacs.
\begin{flushright}
\cite{barchiesi1996review}
\end{flushright}
\end{adjustwidth}

\begin{adjustwidth}{0.07\linewidth}{0.07\linewidth}
\footnotesize%
[I]n the last twenty years scholars have started to read these poems as Ovid's
erotica in exile\dots This is surely right. Stylistically the double epistles
belong among the exile poetry, and with the later poems rather than the early
\emph{Tristia}.
\begin{flushright}
\cite[144--5]{heyworth_16}
\end{flushright}
\end{adjustwidth}

This simple assumption also obviates another objection---that the letters are
not mentioned in \emph{Amores} 2.18 or indeed anywhere else in Ovid's work
(unusual for a poet who loved to advertise). Hence the authorship of these
poems is inextricably entwined with their dating. If they are shown to be
Ovidian then they can only be late works, and if they are shown to be late it
will be much easier for critics to accept them as Ovidian. I believe that the
stylometric analysis demonstrates both.

\section{Methodology}
\label{sec: methodology}

To begin, there are one or two important points that should be made regarding
methodology. As mentioned above, the standard approach in authorship questions
has been to proceed negatively, by identifying stylistic anomalies. These
often take the form of `X never uses this verb' or `the word Y does not appear
in poetry'---in other words the focus is on \emph{rare events}. To some extent
this is a valuable function of the human brain, which is extraordinarily good
at noticing contextual anomalies in complex systems. As a stylometric
approach, however, it is dangerous. The other methodological disease is the
intrusion of subjective judgement, of which two common symptoms are abuses of
`Axelson's Law' in questions of priority and the conflation of `quality' with
`genuineness'. Although these kinds of approaches are far from a thing of the
past, the latter half of the last century began a slow movement away from
them. \citeA[84]{wilkinson_74} observes that

\begin{adjustwidth}{0.07\linewidth}{0.07\linewidth}
\footnotesize%
the nineteenth century laid arrogant hands on the works of many authors. Large
portions of the traditional texts were bracketed as interpolations simply
because a particular editor felt them to be unworthy of the author, subsidiary
`arguments' being easily discovered \emph{a posteriori}. The \emph{Heroides}
have received much attention of this kind.
\end{adjustwidth}

\citeA{kenney1999anomaly}, too, begins an article on the Double
\emph{Heroides} with an admonition to ``beware of hypercriticism'', noting in
particular that ``[s]ingularity is not in itself a ground for suspicion. It
makes no sense to require that a writer shall never do anything unless he does
it at least twice.'' From a statistical point of view, the problem with
focusing on rare events is \emph{variance}. If I knew the average number of
passengers on a London bus, I would find many `anomalies' as I wandered the
city---buses completely empty, or packed full. However, if I knew the average
number of commuters for the entire city on Mondays, that number would be much
more reliable as a benchmark for unusual activity. This, then, is the idea
behind multivariate analysis, and the reason to focus not on rare events and
exceptions but on stylistic events that happen in every line.

It will sound, at this point, as if I am about to proclaim the death of human
criticism and the ascendency of the machine, but nothing could be further from
the truth. In the first place, computational methods are most useful to test
hypotheses, and those hypotheses come from the careful literary judgement of
human scholars, once again using that marvellous ability for contextual
analysis. Computational stylometry, like anything else, is capable of
misapplication, misinterpretation or mistake. The great benefit of
multivariate analysis is not to replace scholarly intuition, but simply to
offer an objective tool that can work with thousands, or tens of thousands of
stylistic factors to provide new kinds of evidence to existing areas of
enquiry; instead of looking for anomalies (a negative approach) this is a
\emph{positive} approach, assessing overall conformance to a wide range of
stylistic preferences.

On, then, to technical matters. In this article I perform two kinds of
analysis, poetic and lexico-grammatical. In each case, there are two phases.
The first thing that must be done is to establish that the features are
stylistically discriminatory, in other words that they can be used to tell
authors and works apart. This is evaluated using cross-validated
classification accuracy under several different classification algorithms (to
ensure that the features are generally useful, not just suited to one kind of
model). Note that the goal is not to optimise hyperparameters and find the
most accurate classifier possible, it is simply to understand how well works
and authors separate in these feature universes. Once the validity of the
feature sets has been established, the works are treated as points in high
dimensional space and the relationships between them are interrogated using
cluster analysis, in the context of a broad elegiac corpus. Here, again,
multiple methods are used to ensure that the results are not a fluke of a
single algorithm. I use two different methods to project high dimensional data
to two dimensional figures, UMAP \cite{mcinnes_umap_2018} and t-SNE
\cite{van2008visualizing}. Then, following the general techniques introduced
by \citeA[57--9]{eder2017visualization}, I create a `bootstrap consensus tree'
by aggregating the k-nearest neighbours data for many random feature subsets.
This yields a weighted adjacency graph, which is laid out using
Fruchterman-Reingold \cite{fruchterman1991graph}. In other words two cluster
analyses infer relationships based on high-dimensional \emph{position}, and
one has the positions determined by the strengths of the \emph{relationships}
between works, something like working in the formal mathematical dual.

An important methodological concern is that the units of analysis (i.e. the
poems) are small. Practitioners used to working with modern prose might be
uncomfortable with samples that are smaller than a few thousand words---in our
case, the samples are as small as twenty lines (less than two hundred words).
Based on the classification accuracy, however, there is very good reason to
believe that we can still draw sensible conclusions. Latin poetry is
remarkably stylistically \emph{dense}. A single line of verse is a complex
balance between word choice, metre, ictus, the position of the caesurae, the
effect of dactylic versus spondaic feet, the interplay of sound within the
line and between lines \dots the list goes on; and with each conscious or
unconscious choice the poet leaves stylistic fingerprints. In the
lexico-grammatical analysis, I focus on text \emph{n}-grams, since this too
allows us to extract more information from small samples. For a
morphologically inflected language, a `standard' BOW (bag of words) analysis,
as one might do for English prose, would require that words be lemmatized,
which entails a loss of information. \emph{N}-grams, by contrast, break the
words into smaller chunks (eg `habet' produces the 3-grams `hab', `abe' and
`bet') which captures information both on the lemma as well as on the
grammatical inflection, allowing authorial preferences to be extracted in both
domains. Once again, the classification accuracy shows that this process
extracts enough stylistic signal to reliably separate both authors and works.

Finally, I note that both the complete corpus along with full implementations
of the various analyses are freely available at the associated code repository
\cite{nagy_heroides_2022}.%
    \footnote{The reproduction data is presented in a series of code
    notebooks, using Python scikit-learn \cite{scikit-learn} with
    visualisation via R and ggplot2 \cite{ggplot}.}
\section{The Corpus: `Shorter' Elegy}

\begin{table}
\caption{A summary of the test corpus.}
\label{tab:corpus}
\centering
\begin{tabular}{llrrr}
            &                  &       &   Min.  &  Max. \\
     Author &   Work           & Poems & Length  & Length \\
\midrule
   Catullus &  \emph{Carmina}  &     5 &   24 &  158 \\
       Ovid &  \emph{Amores}   &    49 &   18 &  114 \\
       Ovid &  \emph{Heroides} &    21 &  116 &  376 \\
       Ovid &  \emph{Ex Ponto} &    46 &   22 &  166 \\
       Ovid &  \emph{Tristia}  &    50 &   26 &  578 \\
 Propertius &  \emph{Carmina}  &    91 &    6 &  150 \\
   Tibullus &  \emph{Carmina}  &    16 &   22 &  122 \\
\bottomrule
\end{tabular}
\end{table}

Although this analysis is focused on Ovid's \emph{Heroides}, the validation
phase requires them to be in the broader context of classical elegy, to
confirm that there is a stylistic distinction between both authors and
collections. The test corpus is only moderate in size. It includes 278 poems
by four authors, drawn from seven `works' (which here refers to a collection of
poems, possibly including several `books'). The total size of the corpus is
18,726 lines. I included a few works by Catullus (who wrote little elegy)
since he seemed a useful point of comparison to the Augustan poets, but it did
not seem sensible to venture as far forward as Martial, whose epigrams are a
completely different genre, and mostly just a few lines long. Also excluded
are Ovid's `long' works, \emph{Ars Amatoria}, \emph{Fasti}, \emph{Remedia
Amoris} etc.---not only do we have enough Ovid already but there are also
genre concerns. The final selection is summarised in Table \ref{tab:corpus}.
Finally, the text of every poem is phonetically transformed. This is needed in
any case to perform the rhyme analysis, but it also enables the \emph{n}-gram
statistics to more accurately isolate sonic preferences---for example to
detect an authorial preference for a /k/ sound beginning a word we would
otherwise need to consider orthographic `c' and `qu' (Fig.
\ref{fig:phon_trans}).

\begin{figure}
\caption{The start of the \emph{ES}, before and after phonetic transformation.}
\label{fig:phon_trans}
\begin{adjustwidth}{0.07\linewidth}{0.07\linewidth}
ecquid ut inspecta est studiosae litera dextrae\\
\phantom{xx}protinus est oculis cognita nostra tuis\\
an nisi legisses auctoris nomina sapphus\\
\phantom{xx}hoc breue nescires unde ueniret opus\\
\smallskip

ekkwid ut inspekta\_st studiosae litera dekstrae\\
\phantom{xx}protinus est okulis konjita nostra tuis\\
an nisi legisses auktoris nomina sappus\\
\phantom{xx}hok brewe neskires unde weniret opus\\
\end{adjustwidth}
\end{figure}

\section{Results: Lexico-Grammatical Analysis}
\label{sec: lsa}

\subsection{Limitations and General Accuracy}
\label{sec: ngram_acc}

\begin{figure*}
  \caption{How does the classification accuracy change as the minimum poem
  size in the corpus increases? A comparison using four different algorithms.}
  \label{fig:ngram_acc}
  \centering
  \subfigure[Accuracy by Author]{
    \includegraphics[width=0.45\textwidth]
    {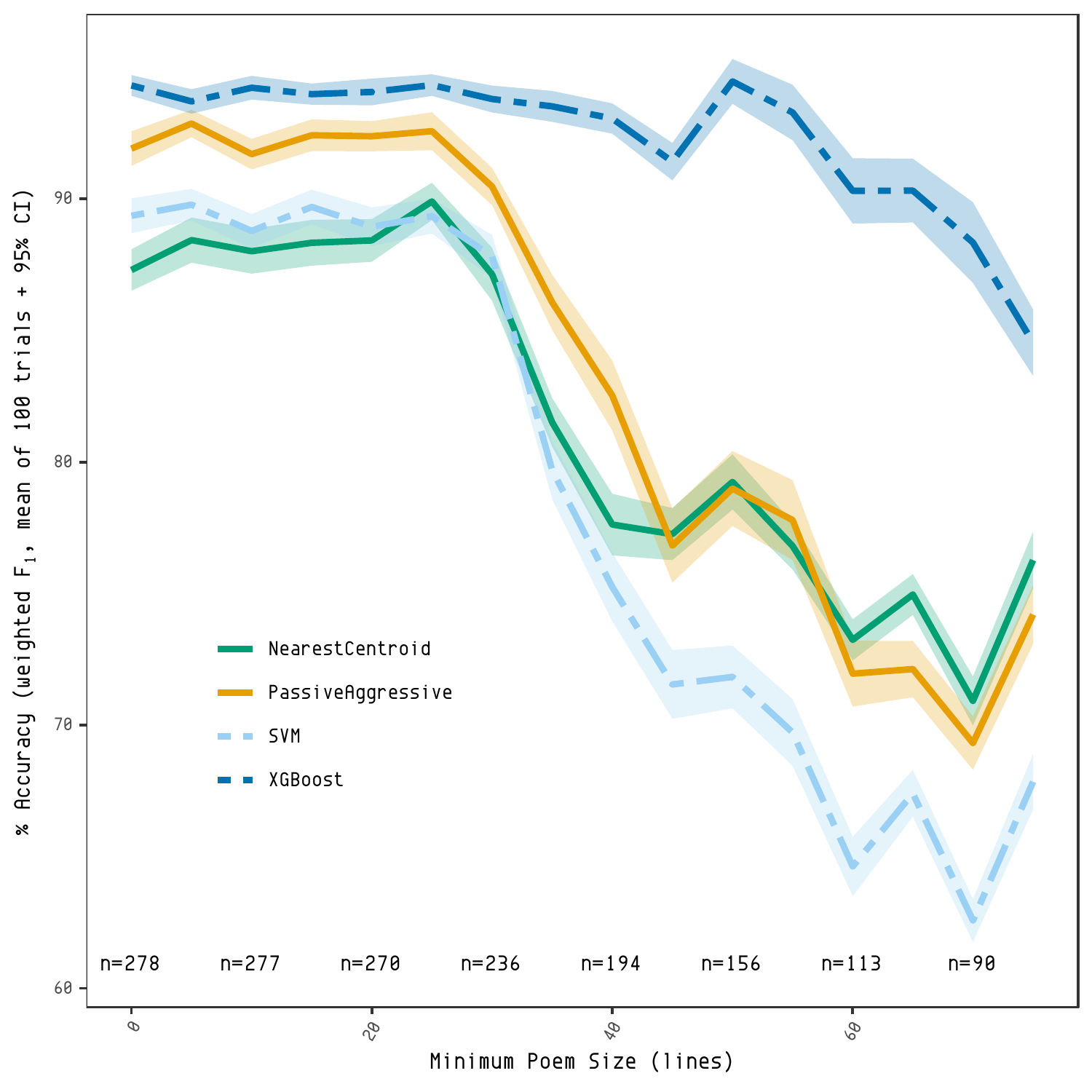}
  }
  \qquad
  \subfigure[Accuracy by Work]{
    \includegraphics[width=0.45\textwidth]
    {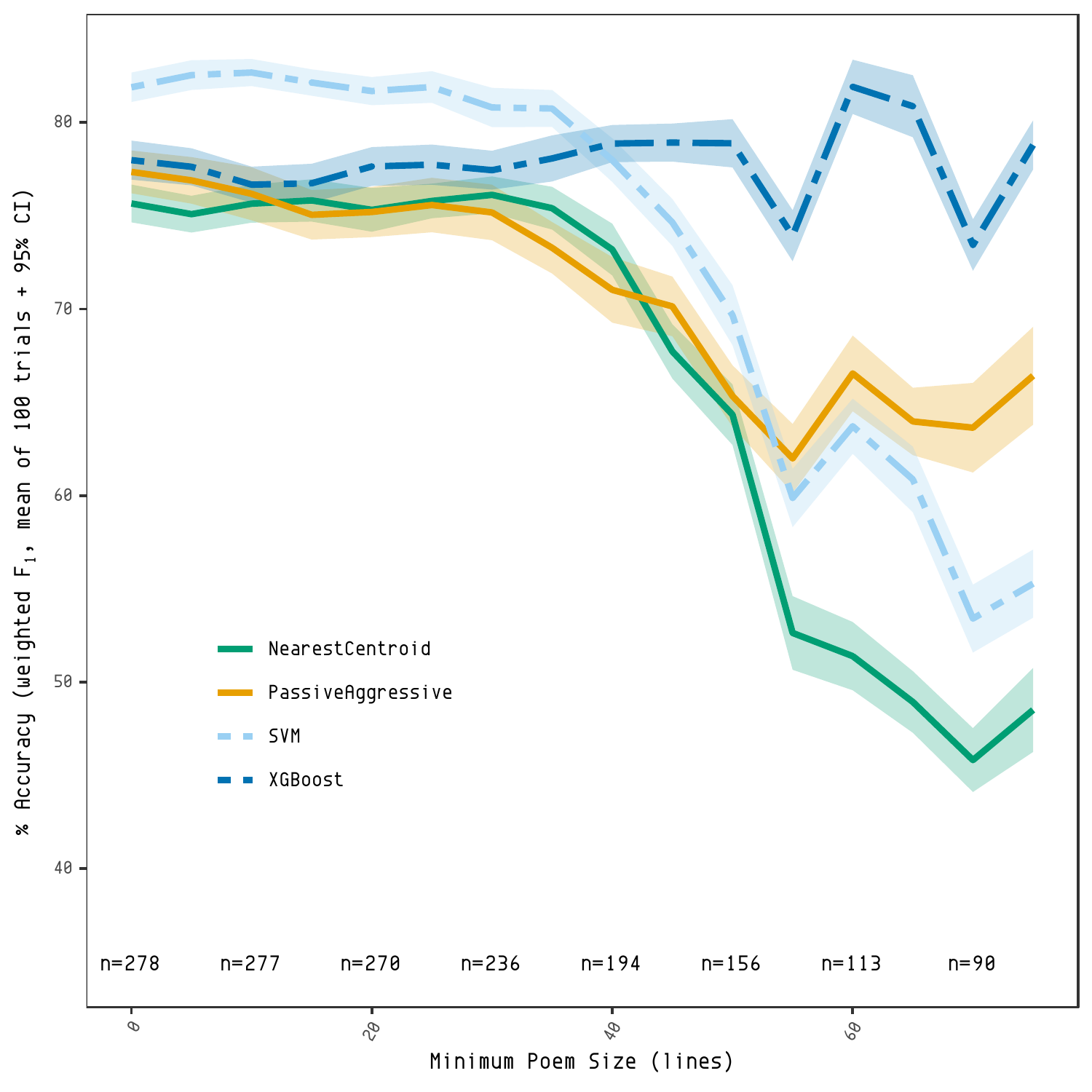}
  }
\end{figure*}

As outlined above, the lexico-grammatical analysis is conducted using TF-IDF
scaled frequency counts for character 2-, 3-, and 4-grams. This results in a
high dimensional feature space (about 30,000 features); in such environments
some care needs to be taken. For hugely overspecified systems, the `curse of
dimensionality' may manifest in a variety of different ways
\cite{zimek_etal}---in this case, some classification models perform poorly,
and dimension-reduction methods like t-SNE are unreliable unless the data is
first reduced by some other means. To avoid this, the \emph{n}-gram data is
first reduced to 50 dimensions using Singular Value Decomposition. This
technique, when applied to TF-IDF data, is fairly standard and is called
Latent Semantic Analysis or LSA. Another potential source of error is that the
results are affected by topic---for example poems about farming will employ
distinctive agricultural lexicon, which can mask or overpower lexical choices
imposed by authorial style. Nevertheless, character \emph{n}-grams are a
powerful and well-attested technique, shown here to effectively distinguish
both authors and works.

\begin{figure}
\caption{
  Confusion matrix, mean of 100 trials. Entries show the percentage of times
that a y-axis work was classified as the given x-axis work. Classifier is
scikit-learn \texttt{NearestCentroid()} using an 80/20 test/train split.
Training data is TF-IDF transformed 2-, 3- and 4-gram frequencies for each
work, reduced to 50 dimensions with SVD. }
\label{fig:cm_ngrams}
\includegraphics[width=0.45\textwidth]{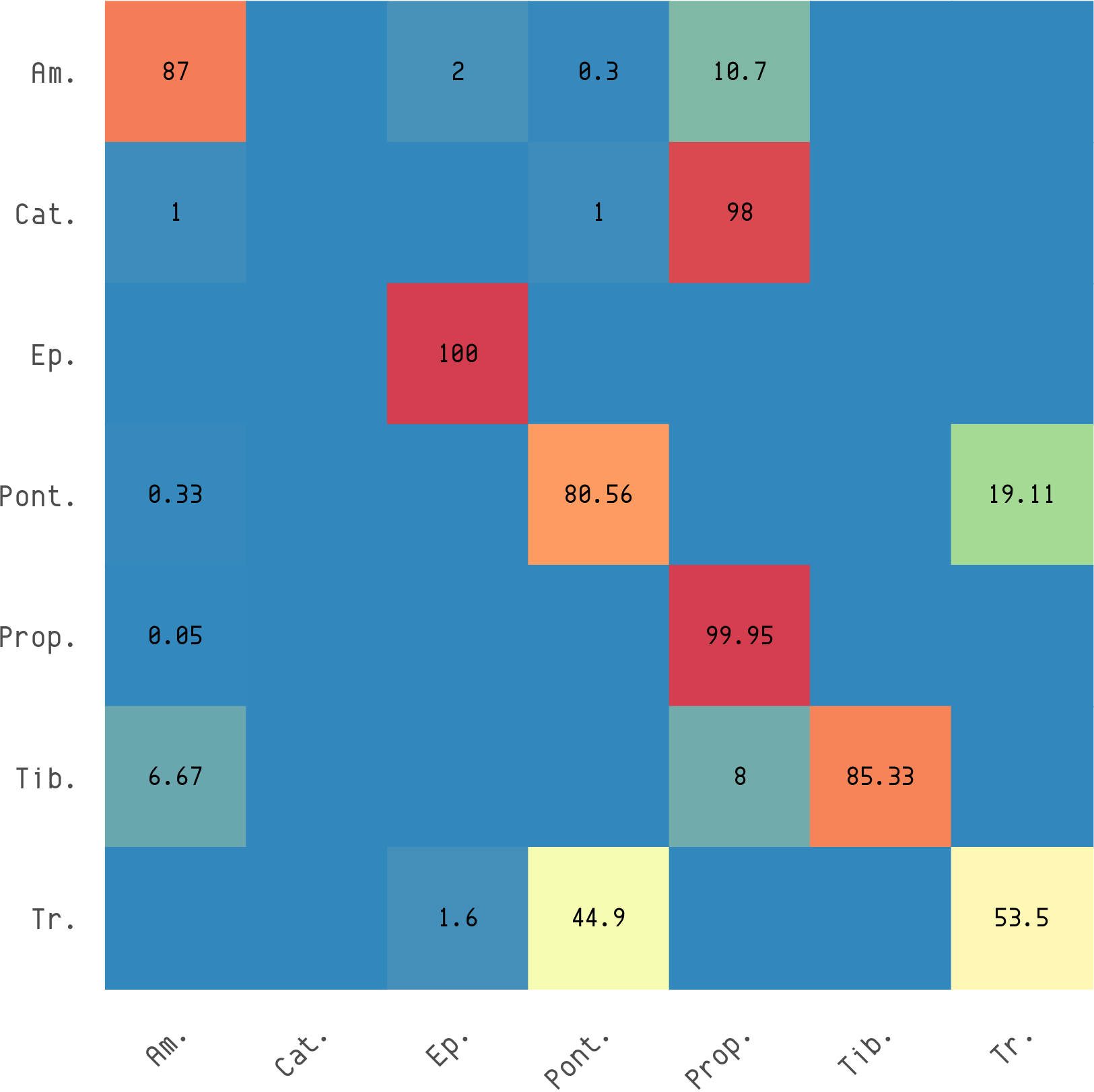}
\end{figure}

First we consider the general accuracy of the supervised classification
algorithms. As mentioned, this is presented as a proxy for the general utility
of the feature universe; a way of estimating how well authorial style is
reflected in these features. Since the size of the poems is a potential cause
for concern, Fig. \ref{fig:ngram_acc} examines the classification accuracy by
both work and by author, when considering poems longer than a certain
size---the hypothesis being that accuracy should improve if the analysis is
restricted to larger poems. As can be seen, this is not the case. In this
instance it appears that increasing the size threshold lowers the number of
poems available for training, damaging the overall accuracy. Thus, a
conservative minimum size of 20 lines was chosen for the cluster analysis,
eliminating only eight poems. Next, Fig. \ref{fig:cm_ngrams} shows the
confusion matrix for the \texttt{NearestCentroid} classifier (by Work). This
shows which works are consistently well-classified over many test-train
splits, implying that they are fairly distinctive. As can be seen, the
\emph{Heroides} (Ep.) are perfectly classified; they can be reliably be
distinguished from Ovid's other works as well as from the works of other
authors. It is also apparent from the confusion matrix that the main source of
inaccuracy in the model is the stylistic similarity between Ovid's
\emph{Tristia} (Tr.) and \emph{Ex Ponto} (Pont.) (this is to be
expected---both works are exilic and share many themes) and so the `true'
accuracy is somewhat better than the numbers in Fig. \ref{fig:ngram_acc} would
suggest.%
    \footnote{Eliminating just \emph{Ex Ponto} from the corpus increases
    per-work accuracy by about 10\% for both the LSA and poetic models.}

\subsection{Cluster Analysis}

\begin{figure*}
  \caption{Cluster analysis of the LSA data (dimension=50), showing
  lexico-grammatical style.}
  \label{fig:ngram_cluster}
  \centering
  \subfigure[Bootstrap Consensus Tree. Aggregated kNN (k=3, metric=cosine)
  data from 500 subsets of 15 features. Layout via Fruchterman-Reingold,
  thicker edges are stronger links.]{
    \includegraphics[height=0.5\textheight,keepaspectratio]
    {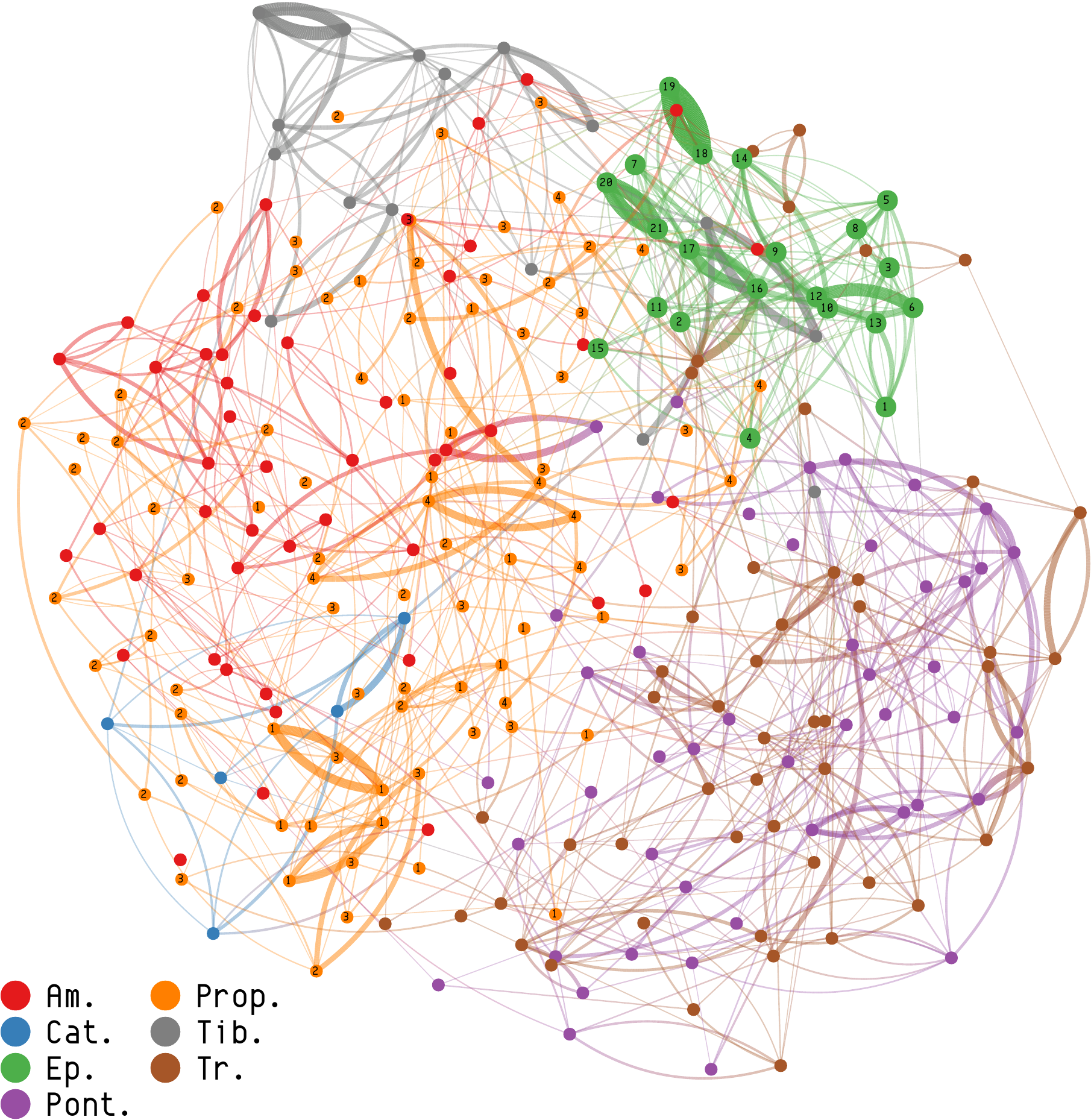}
  }
  \subfigure[UMAP Projection (metric=euclidean)]{
    \includegraphics[width=0.48\textwidth,keepaspectratio]
    {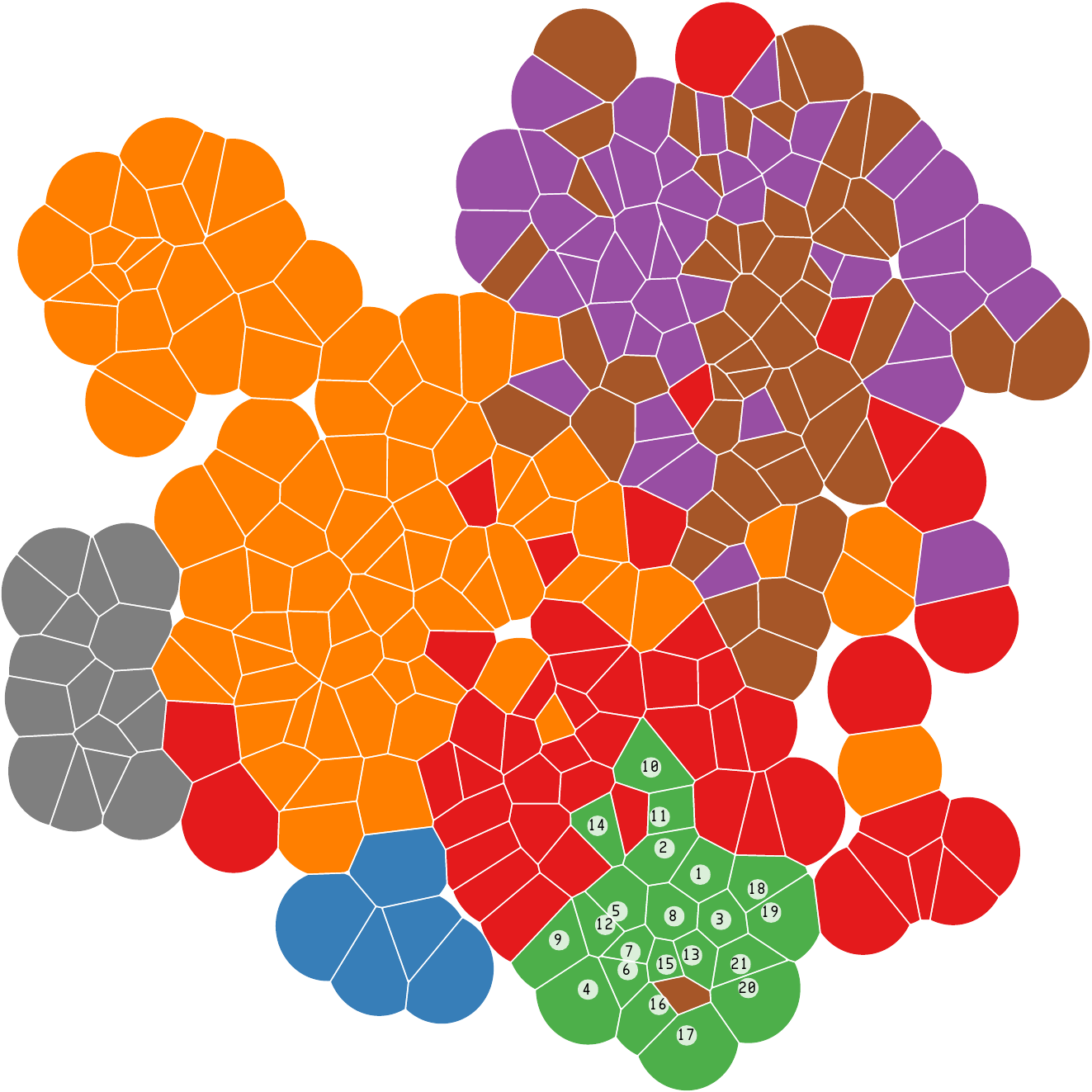}
  }
  \subfigure[t-SNE Projection (metric=euclidean, perplexity=10)]{
    \includegraphics[width=0.48\textwidth,keepaspectratio]
    {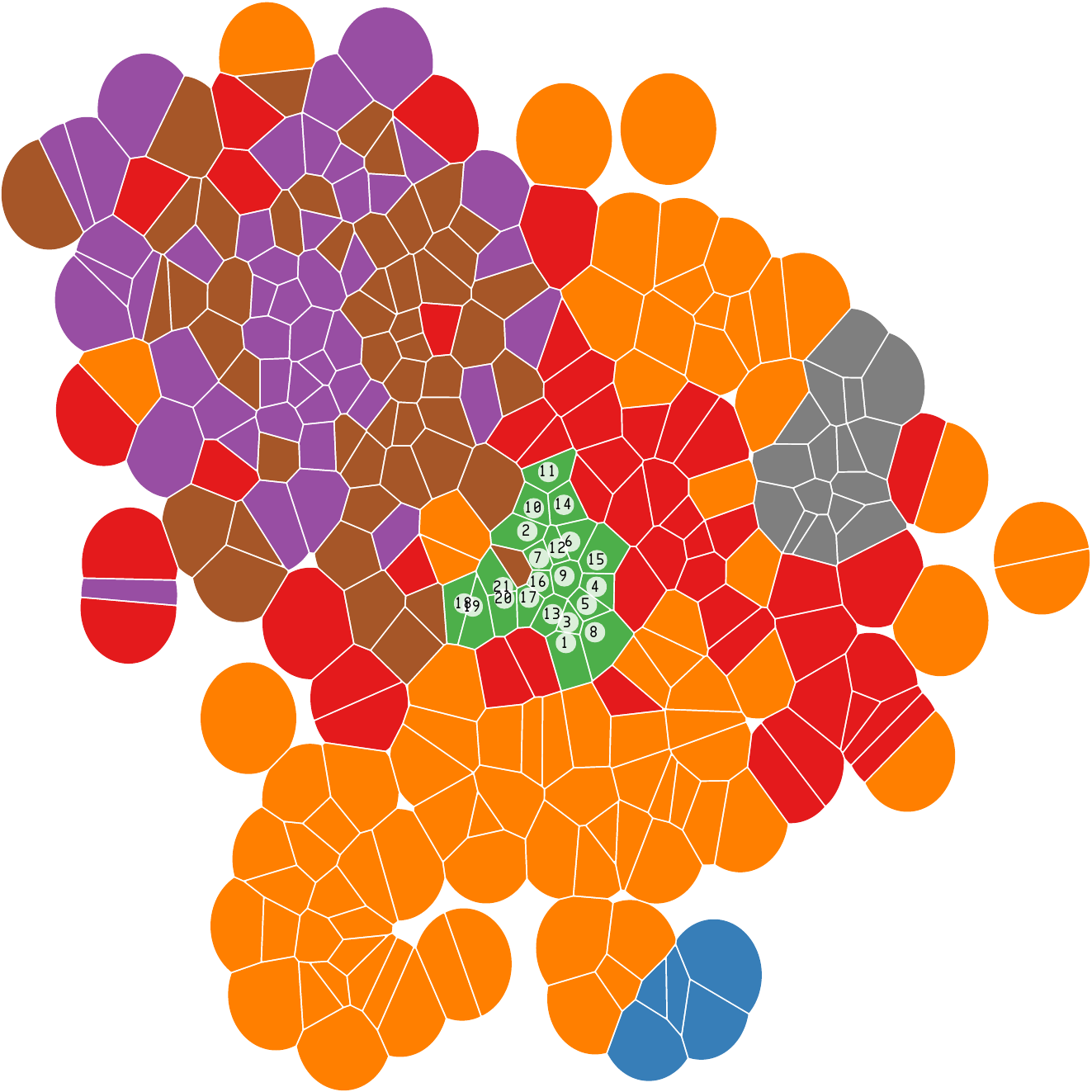}
  }
\end{figure*}

The previous section established that \emph{n}-gram frequencies for the poems
contain enough stylistic signal to discriminate meaningfully between both
authors and works. Given that this is the case, it is now reasonable to
examine the results of unsupervised cluster analysis. The general assumption
for all clustering algorithms is that points which are `close' are somehow
related, however the underlying mathematics (as well as precisely what is
meant by `close') can vary greatly. Here, three different clustering
algorithms are used, and it is reassuring that the general picture is the same
in all three cases. In Fig. \ref{fig:ngram_cluster} we are looking at a latent
semantic analysis, which includes a great deal of information from the
specific lexicon being employed (from which, if the goals were different, the
semantic field could be inferred). Unsurprisingly, in Fig.
\ref{fig:ngram_cluster}(a) we see a great deal of semantic overlap between
Ovid's \emph{Amores} and the works of Propertius (`canonical' love elegy)%
    \footnote{The \emph{Amores} are also thought by many to contain deliberate
    imitation of Propertius; as a starting point see \citeA{morgan1977ovid}.}
although Tibullus maintains some stylistic separation. Ovid's \emph{Tristia}
and \emph{Ex Ponto} mostly form their own cluster, and we can see that the
\emph{Heroides} are stylistically distinct, with very strong internal links,
most particularly in the paired Double \emph{Heroides}; 16/17, 18/19 and
20/21. Letter 15, the \emph{ES}, clusters neatly with the rest of the
\emph{Heroides}, giving no cause to question its authenticity. A more or less
identical picture arises from the two other algorithms, Figs
\ref{fig:ngram_cluster}(b) and \ref{fig:ngram_cluster}(c)---The \emph{Amores}
mix with Propertius, Tibullus is a little more distinctive, and the
\emph{Heroides} form a tight group.

\section{Results: Poetic Analysis}
\label{sec: poetic}

\begin{table*}
\caption{A summary of the poetic features}
\label{tab:poet_feats}
\par\medskip
\centering
\begin{tabularx}{\linewidth}{lX}
Feature & Description \\
\midrule
\texttt{HnSP, PnSP} & Proportion of couplets where foot \emph{n} is a Spondee in either the (H)exameter or (P)entameter.\\
\texttt{HnCF, PnCF} & Proportion of couplets with Ictus/Accent Conflict in foot \emph{n}.\\
\texttt{HnDI, PnDI} & \textellipsis Diaeresis in foot \emph{n} (\metricsymbols{_ u u ||} or \metricsymbols{_ _ ||}).\\
\texttt{HnSC, PnSC} & \textellipsis Strong Caesura (\metricsymbols{_ ||}).\\
\texttt{HnWC, PnWC} & \textellipsis Weak Caesura (May only exist in dactylic feet: \metricsymbols{_ u ||}).\\
\texttt{ELC} & Average number of elisions per line. Does not include prodelision.\\
\texttt{LEN} & The length, in lines, of the entire poem.\\
\texttt{RS} & The average rhyme strength. Measures both the number and strength of rhymes (or other sonic correspondences) whether vertical or horizontal.\\
\texttt{LEO} & The average number of `leonine' rhymes (rhymes between the words at the central caesura and the end-of-line).\\
\texttt{PFSD} & The standard deviation of the length (in syllables) of the final word in the pentameter. Early Ovidian practice (ending every pentameter with a disyllable) would show a \texttt{PFSD} of zero.\\
\bottomrule
\end{tabularx}
\end{table*}

In this section I shift from well-established methods in computational
stylometry to newer techniques that examine non-lexical features that are
indigenous to poetry. This combines and extends research that I began on
hexameter metre \cite{nagy2021metre} as well as work on the stylistic signal
contained in deliberate sonic correspondence \cite{nagy_rhyme_2022}. Thus,
while the results are emphatic and the techniques appear powerful, it should
be made clear that this is an area of emerging research. The features being
examined here are purely poetic---there is no consideration of the words being
used, only the way in which the verse is constructed. The features are
separated into broad domains. First, foot patterns and pauses, which includes
the type of foot at each position (dactyl or spondee) as well as the position
of all caesurae and diaereses; second the interplay between ictus (the start
of the metrical foot) and accent (syllables bearing stress); and third, the
sonic style, including the number and strength of `sonic correspondences'
(rhyme, roughly speaking) of various kinds. Finally, since it is crucial to
the traditional debate, I measure the degree to which the length of the final
word in the pentameter is allowed to vary, and also consider the overall
length of the poems (this was included because some of the \emph{Heroides}, 16
in particular, are considerably longer than the general trend; a point which
has been raised during debate over its authenticity). These features are
summarised in Table \ref{tab:poet_feats}. Not every possible feature is
measured---for example the final two feet of the pentameter are mandatory
dactyls, and so the features \texttt{P3SP} and \texttt{P4SP} would contain no
stylistic information. The final universe contains 43 features; a reasonable
number, but hopefully few enough to escape the label of `high dimensionality'.
The hope (which seems to be borne out by the results) is that poetic features
are less obscured by things like topic and genre, and should present a cleaner
stylistic signal.

\subsection{Limitations and General Accuracy}

\begin{figure*}
  \caption{How does the classification accuracy change as the minimum poem
  size in the corpus increases? A comparison using four different algorithms.}
  \label{fig:poetics_acc}
  \centering
  \subfigure[Accuracy by Author]{
    \includegraphics[width=0.45\textwidth]
    {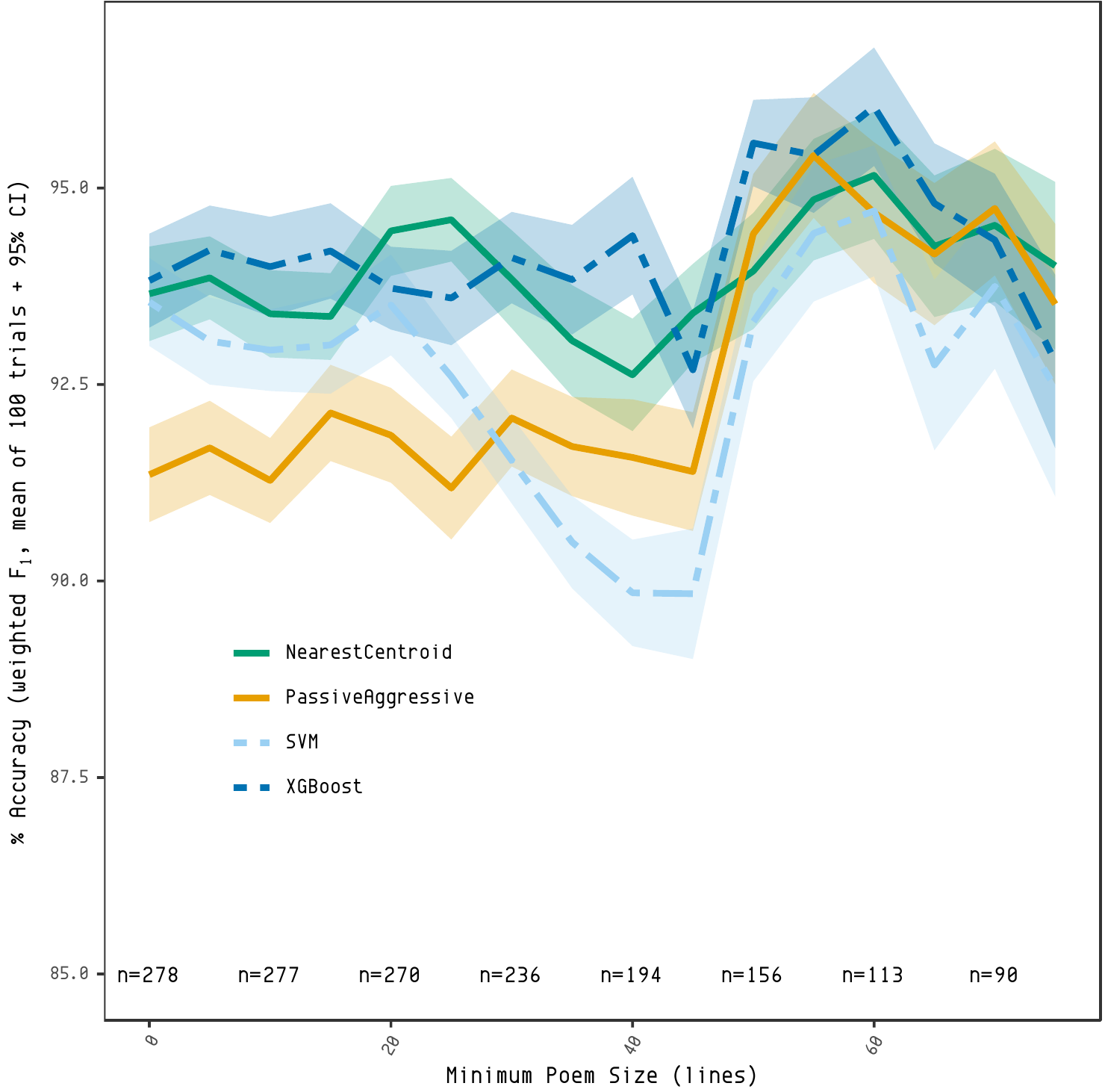}
  }
  \qquad
  \subfigure[Accuracy by Work]{
    \includegraphics[width=0.45\textwidth]
    {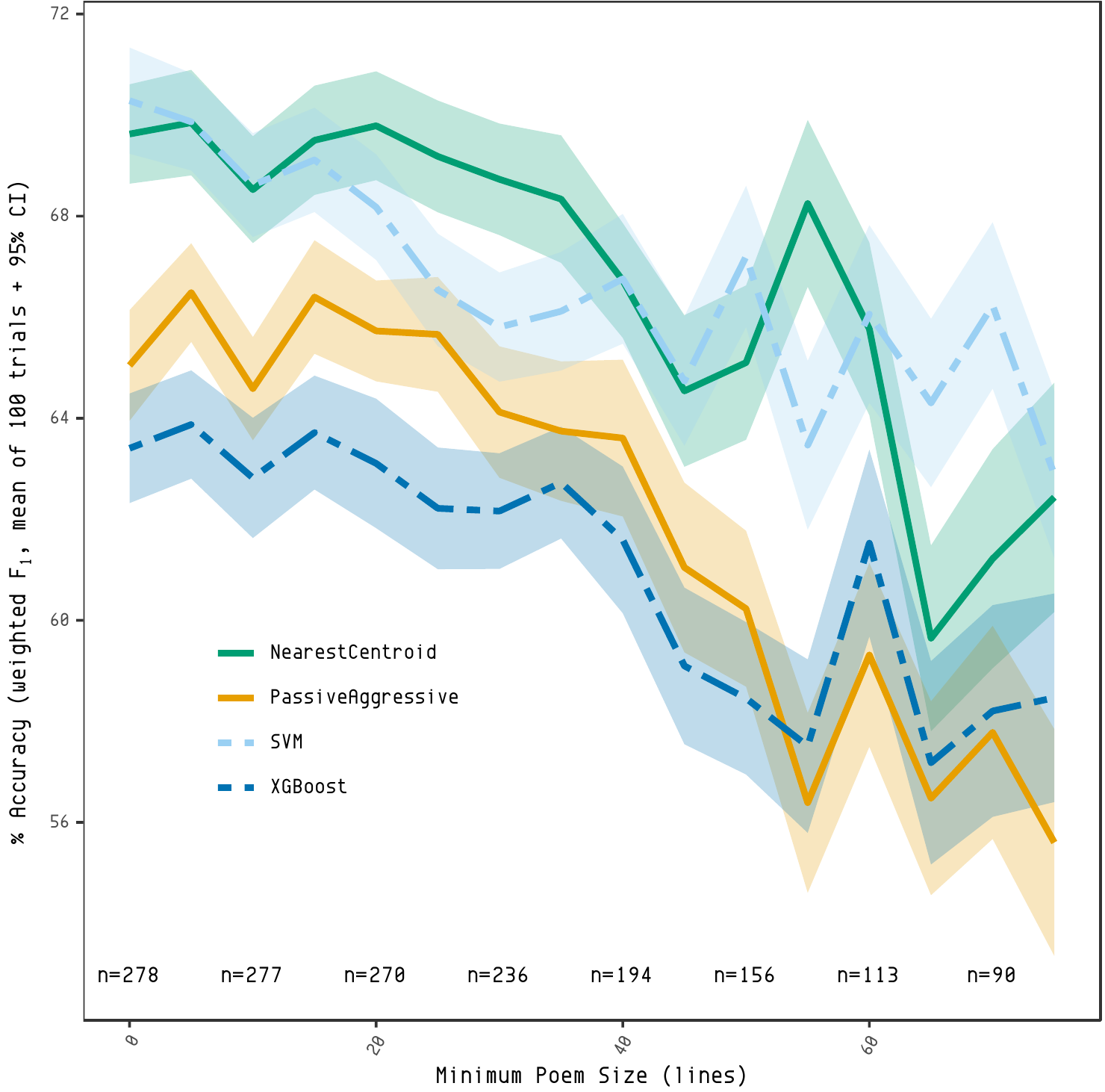}
  }
\end{figure*}

\begin{figure}
\caption{
  Confusion matrix, mean of 100 trials. Entries show the percentage of times
that a y-axis work was classified as the given x-axis work. Classifier is
scikit-learn \texttt{NearestCentroid()} using an 80/20 test/train split.}
\label{fig:cm_poetics}
\includegraphics[width=0.45\textwidth]{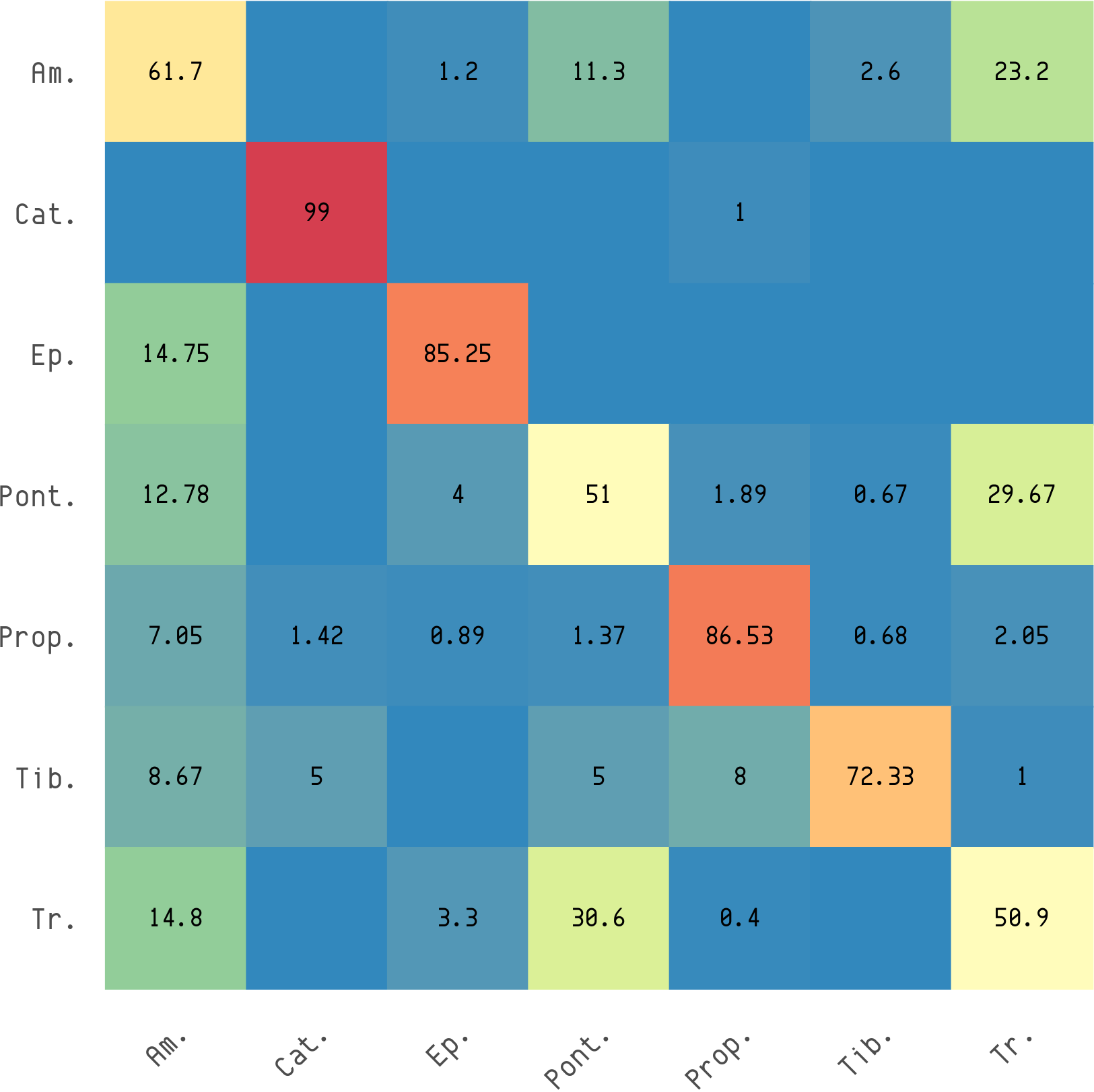}
\end{figure}

As before, the first task is to establish that the features are able to
meaningfully distinguish authorial style. The results here are broadly similar
to those in \S \ref{sec: ngram_acc}. In Fig. \ref{fig:poetics_acc} we see
again that the accuracy decreases as the shorter poems are filtered out. The
best performing classifier is now \texttt{NearestCentroid}, and the F$_1$
accuracy by Work (the most important figure in this context) is a little
lower---although it should hardly be surprising that it becomes more difficult
to identify the theme of a poem if one is forbidden from reading the actual
words! It is important to note, however, that the accuracy by Author is
slightly higher in the poetic models than the previous LSA models. At the risk
of subjectivity, this seems to suggest that the underlying performance of the
models that use poetic features is just as strong, but that the additional
information derived from lexicon and topic in the LSA models gives the
per-Work accuracy an `unfair' boost. In the confusion matrix (Fig.
\ref{fig:cm_poetics}) we see again that the \emph{Heroides} are quite
distinctive in their style, while Ovid's \emph{Tristia} and \emph{Ex Ponto}
are easily confused. Overall, particularly considering the extremely good
performance when identifying authors, the poetic features seem at least as
useful and, based on the clustering results, are probably more so.

\subsection{Feature-Wise Analysis}

In \citeA[1007--9]{nagy2021metre} I discussed a statistical technique that
used the Mahalanobis distance to not only determine how consistent an
observation was with a given comparison set,%
    \footnote{This is a form of outlier detection, sometimes called a
    `one-class problem' in authorship attribution circles, and it has its own
    branch of the literature. \citeA{koppel_fundamental_2012} is a reasonable
    place to start.}
but also to provide a per-feature analysis of the most unusual features. In
this section, the same technique is applied to the elegiac corpus. The
advantage of the Mahalanobis distance is that it corrects for feature
correlation and covariance. In the case of prosodic features there are many
correlations---as just one example, the practice of ending a pentameter with a
disyllable (which primarily affects the feature \texttt{PFSD}) also forces a
weak caesura in the fourth foot (\texttt{P4WC}). Taking the centroid of all
the Ovidian works (164 poems), the Mahalanobis distance was calculated for
every poem in the corpus, representing the degree to which the poems conform
to Ovid's `typical' style. From there, since the (squared) Mahalanobis
distance is distributed according to $\chi^2$, a \emph{P}-value can be
calculated. The method is reasonably powerful. Of the (102) non-Ovidian works,
only five would be accepted as Ovidian style at the 99\% confidence level.%
    \footnote{In order of decreasing similarity: Prop. 4.11, 4.6, 3.24, 4.4;
    Tib. 1.4.}
Of the (164) Ovidian works, there are fourteen outliers at that same
confidence level,%
    \footnote{Mostly from \emph{Tristia} and \emph{Ex Ponto}, but also
    \emph{Amores} 1.11, 2.8, and 2.13.}
but none of the \emph{Heroides}, including, obviously, the \emph{ES}, are
outliers at any level of confidence. The obvious objection to this analysis is
that none of the non-Ovidian poems were attempts at imitation. Could an
interpolator have simultaneously imitated every one of Ovid's poetic
tendencies (while also using Ovidian lexico-grammatical style)? In my view,
no; but Late Antique and Humanist authors achieved levels of Latinity that few
today could match. Or, could the statistical features (foot patterns, caesura
positions, etc.) be imitated `instinctively' by a gifted poet who had
completely internalised Ovidian style? In the end, it cannot be ruled out.
However, as previously pointed out, we would be dealing with someone with,
essentially, Ovid's talent, as well as an extraordinary gift for imitation.

\subsection{Cluster Analysis}

\begin{figure*}
  \caption{Cluster analysis of the z-scaled data (43 features), showing poetic style.}
  \label{fig:metre_cluster}
  \centering
  \subfigure[Bootstrap Consensus Tree. Aggregated kNN (k=3, metric=cosine)
  data from 500 subsets of 15 features. Layout via Fruchterman-Reingold,
  thicker edges are stronger links. \emph{Heroides} enlarged and numbered,
  works by Propertius numbered by Book.]{
    \includegraphics[height=0.48\textheight,keepaspectratio]
    {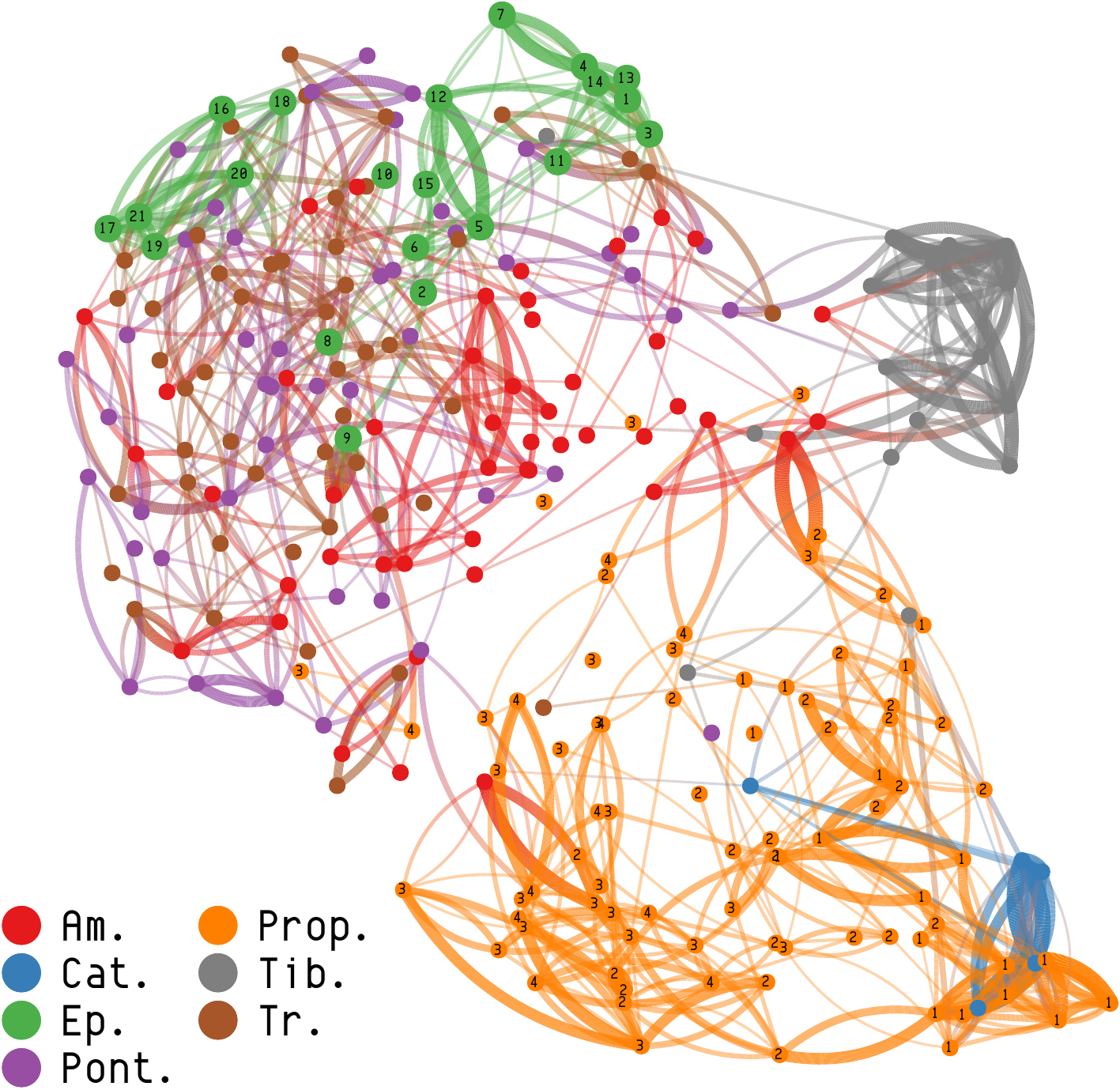}
  }
  \subfigure[UMAP Projection (metric=euclidean)]{
    \includegraphics[width=0.48\textwidth,keepaspectratio]
    {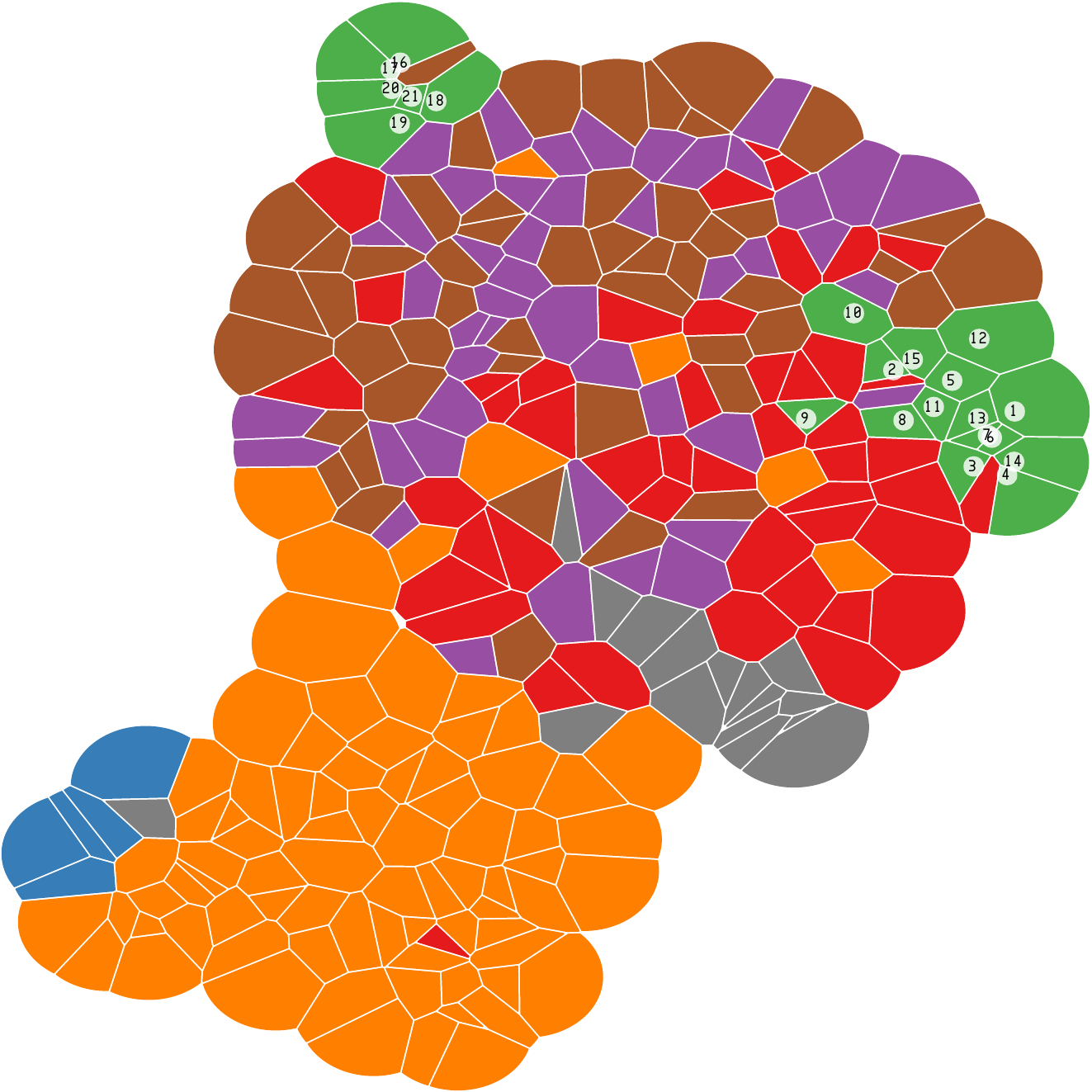}
  }
  \subfigure[t-SNE Projection (metric=euclidean, perplexity=12)]{
    \includegraphics[width=0.48\textwidth,keepaspectratio]
    {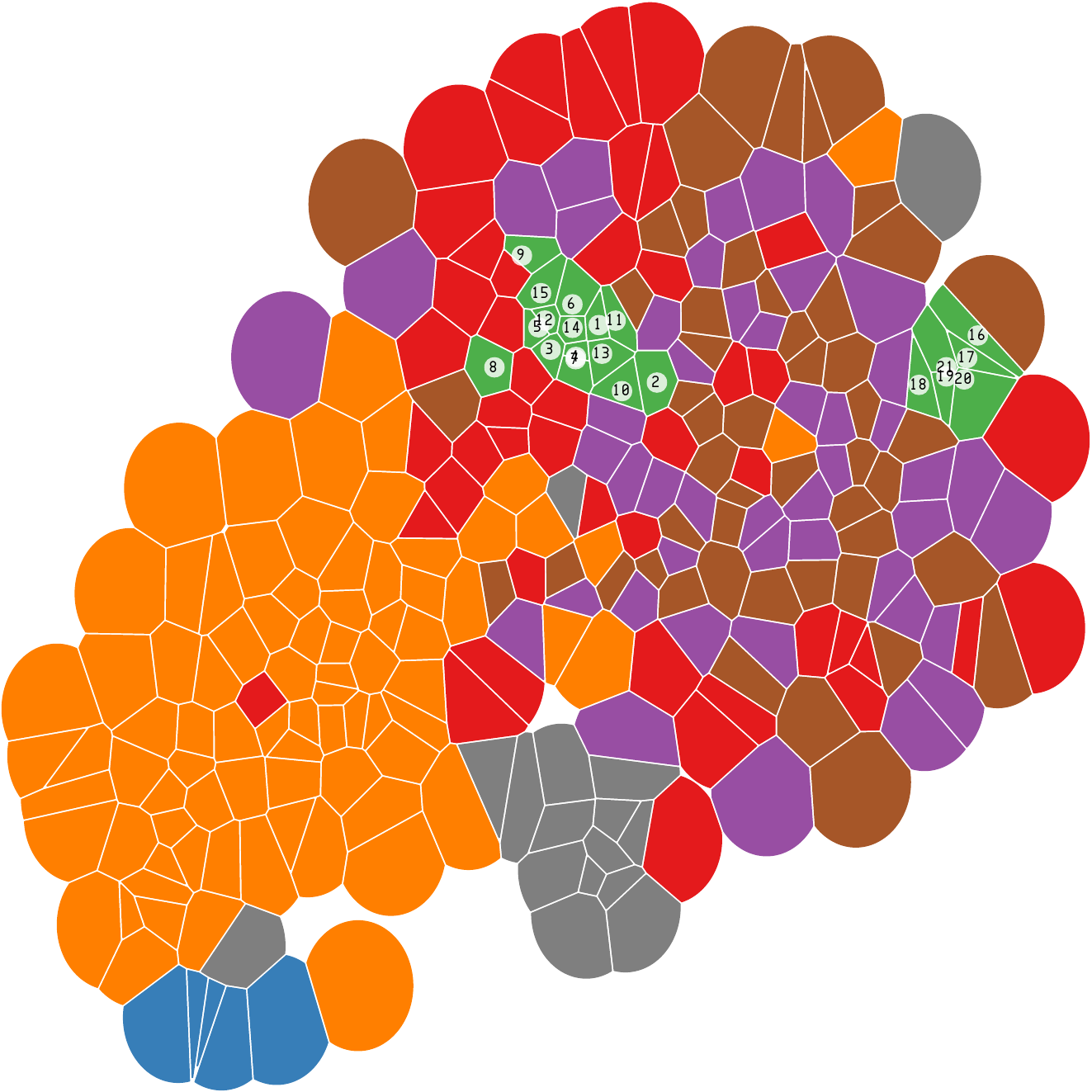}
  }
\end{figure*}

Looking at the clustering results, the first observation is that there is
clear support for the general hypothesis that poetic features provide a
cleaner stylistic signal. Where Fig. \ref{fig:ngram_cluster} suffered from
significant overlap between authors that was driven by topic, in Fig.
\ref{fig:metre_cluster}(a) we have clear separation by authorial style, as
well as further discrimination within Author clusters that appears to be a
temporal signal (discussed in \S \ref{sec: temporal}). In all three analyses,
the bulk of the Single \emph{Heroides} form one cluster within Ovidian style,
and the Double letters cluster tightly among themselves, showing, again, a
stylistic `break' between the two groups. The \emph{ES} again falls in the
middle of the Single letters---it is completely typical of the style of this
group, as well as being characteristically Ovidian. In the UMAP and t-SNE
projections (Figs \ref{fig:metre_cluster}(b) \& \ref{fig:metre_cluster}(c))
the separation between the \emph{Amores} and \emph{Tristia}/\emph{Ex Ponto}
can be seen more clearly, with the Single \emph{Heroides} clustering in with
the former and the Double letters with the latter.

\subsection{Searching for a Temporal Signal}
\label{sec: temporal}

\begin{figure*}
  \caption{Progression of the prosodic style of the \emph{Heroides} showing
  the comparative similarity to either the \emph{Amores} (early style) or
  \emph{Ex Ponto} (late style). In grey: a GAM trendline with 95\% confidence
  interval.}
  \label{fig:ampont}
  \centering
  \subfigure[SVM (relative distance from hyperplanes)]{
    \includegraphics[width=0.45\textwidth]
    {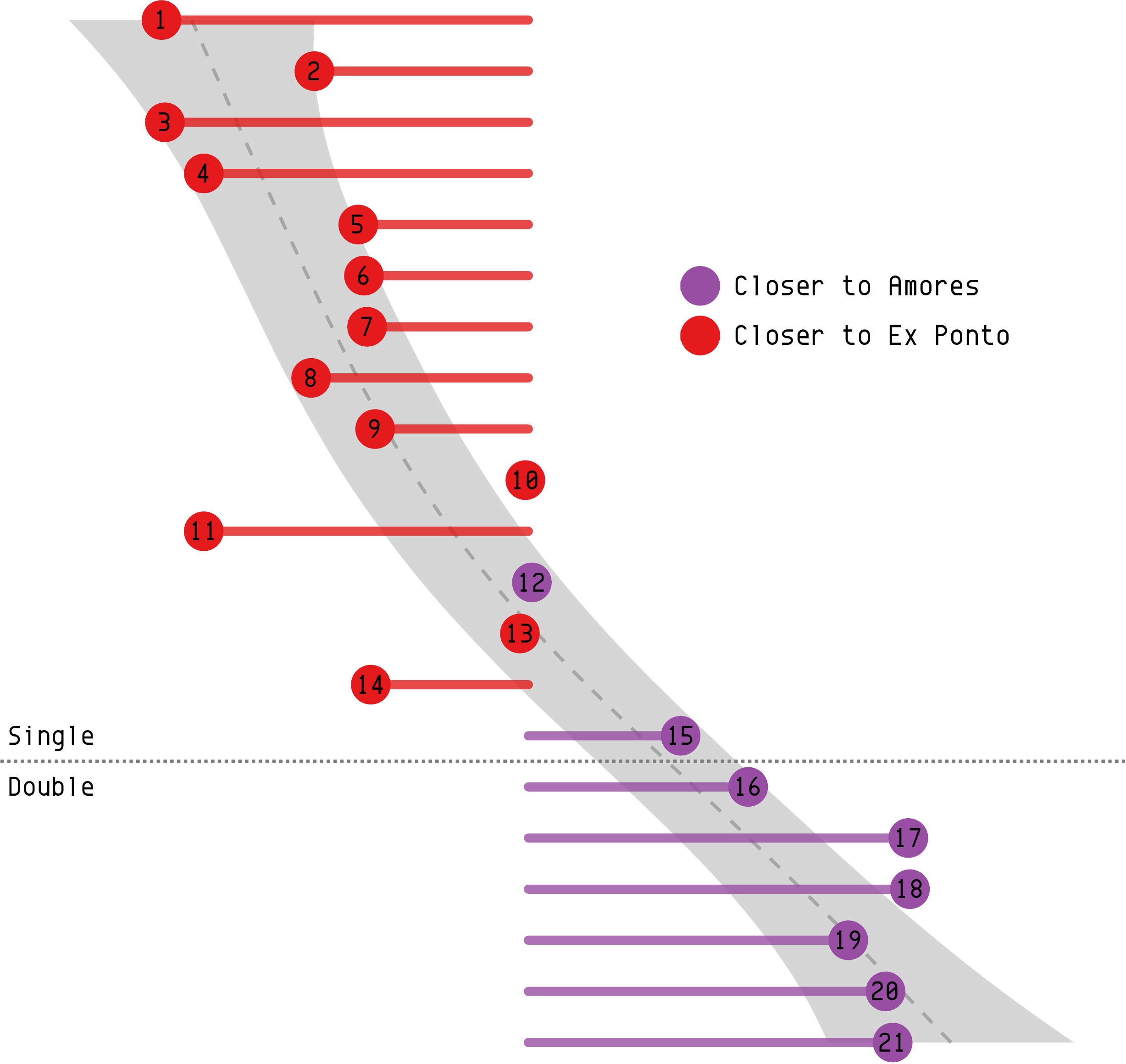}
  }
  \qquad
  \subfigure[Nearest centroid (Euclidean distance)]{
    \includegraphics[width=0.45\textwidth]
    {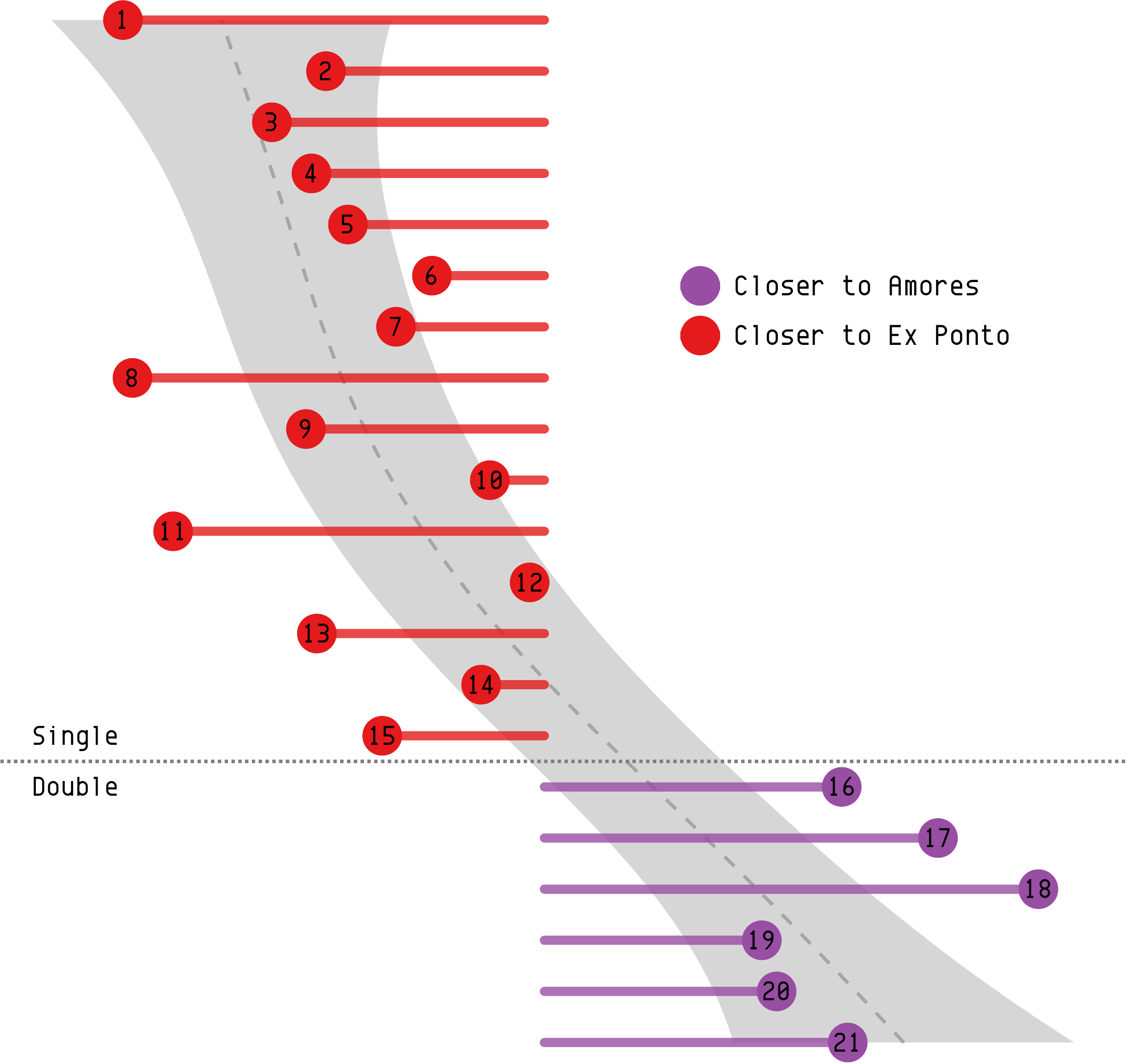}
  }
\end{figure*}

As a final experiment, here is a more specific attempt to detect a temporal
signal in the poetic style (recall that by `poetic' I mean specifically
non-lexical style---prosody, metre, soundplay\textellipsis) of the
\emph{Heroides}. The central assumption is that we have two points of
comparison, the \emph{Amores} for early style, and the \emph{Epistulae Ex
Ponto} for late style. There are complications, though, in the `early' style
of the \emph{Amores}---not only was the collection heavily edited from five
books to three, but modern scholarship tends towards the view that some of the
poems were included or revised later in Ovid's career.%
    \footnote{On these and other chronology questions, the most up-to-date
    summary is \citeA[i--iii]{davis_amores}. The idea that parts of the
    \emph{Amores} may have been revised over decades might account for the
    wide variation in style that can be seen in Fig. \ref{fig:metre_cluster}.}
Nevertheless, these are the best witnesses we have to Ovid's earliest elegiac
style. The general approach taken is to consider each letter and plot the
relative similarity to either the \emph{Amores} or to \emph{Ex Ponto}, with
the centre of the graph being stylistically equidistant. There is no attempt
here to classify the \emph{Heroides}, and no test/train partition of the data.
Two models are trained on the full corpus, one \texttt{SVM} and one
\texttt{NearestCentroid} (these models are the easiest from which to extract a
relative similarity metric, and were also the most generally accurate in
testing---see Fig. \ref{fig:poetics_acc}). In Fig. \ref{fig:ampont} we see the
results. Although the fine details differ according to the model, the results
support two assertions: first that the style of the \emph{Heroides} gradually
evolves from `early' to `late', supporting the idea of sequential composition;
second, that the Double \emph{Heroides} seem to be markedly `later' in style,
supporting the view that they were composed in exile. In terms of the
\emph{ES}, the style seems `fairly late', which is consistent with the other
evidence placing it in its present position as the fifteenth letter, but in my
view the results are too variable to make that case in the absence of other
evidence, and they are certainly too weak to suggest any kind of re-ordering
of the Single letters (for example the eleventh letter from Canace to
Macareus). Since the temporal signal seems genuine, it is tempting to look at
the tight cluster of the Double \emph{Heroides} in Fig.
\ref{fig:metre_cluster}(a) as evidence that they were composed within a fairly
short space of time (since there is little stylistic drift, unlike the Single
letters), but that is merely speculation.

Although it is not directly relevant to the \emph{Heroides}, it is also
interesting to consider the poetic development of Propertius, which can be
seen strongly in Fig. \ref{fig:metre_cluster}(a). The first book (called the
\emph{monobiblos}) is strongly self-similar, and there is a fairly obvious
split between the first two books (mainly Cynthia poems) and the third and
fourth which some argue were subject to the influence of Maecenas (and thus
indirectly of Augustus).%
    \footnote{As well as the potential for political influence, which after
    all might be expected to affect genre and topic but not necessarily
    style, it should be borne in mind that the interaction between the various
    poets of Maecenas' circle was likely to produce some stylistic
    cross-pollination.}
There also appears to be a fairly close link between the \emph{monobiblos} and
the poetics of Catullus, which has a bearing on the stylistic origins of
Augustan elegy.%
    \footnote{Catullus' metrical technique was once considered `careless and
    uncouth', but this view was challenged by D. A. \citeA{west1957metre} and
    revisited by \citeA{duhigg1971elegiac}. Nobody questions the status of
    Catullus as \emph{some} kind of precursor to the Augustan elegists
    (although much ink has been spilled over the details), but given the
    closeness (based on the cluster analysis, in particular Fig.
    \ref{fig:metre_cluster}(a)) it is worth wondering: did Propertius
    deliberately model his earliest metrics on Catullus? Or perhaps does the
    stylistic link pass through the lost elegies of Gallus? Of course it could
    be mere coincidence---a much more rigorous analysis would be needed before
    pressing this case.}
In summary, this technique seems sensitive, even at this early
stage, to very fine gradations of poetic style; not only authorial, but even
to the development of an author's stylistic preferences throughout their
career. There is significant potential here for future work.

\section{Conclusions}

\citeA[395]{kenney1979two} observed that when attempting to prove authenticity
 (as opposed to disputing it) ``[t]he most that it is usually realistic to
 expect is what I have called a verdict of \emph{nihil obstat}: a conclusion
 that, on balance, a disputed work or passage contains nothing fundamentally
 inconsistent with its reputed authorship''. In this article I have attempted
 to show that, with computational methods, it may be possible to expect more.
 More than simply showing that there is nothing inconsistent with Ovidian
 authorship, multivariate analysis demonstrates that the style of the
 \emph{Heroides}, including the Double letters (and including, by the way,
 \emph{Heroides} 12 whose authenticity has at times been questioned) conforms
 closely with dozens of features that reflect Ovid's typical practice. This is
 \emph{positive} evidence---in every poetic feature that was measured, whether
 the distribution of dactyls and spondees in each foot position, whether the
 type and frequency of caesurae, the frequency of elision, even the frequency
 and position of ictus conflicts, in all of these measurements the
 \emph{Heroides} are consistent with Ovidian practice. This does not `prove'
 Ovidian authorship, but it demonstrates the incredible lengths, far beyond
 simply writing Ovidian words, to which the hypothetical interpolator must go
 to produce a false positive. Two kinds of analysis were performed, and in
 each case it was shown that there is enough stylistic signal available to
 reliably differentiate both authors and works. LSA (\S \ref{sec: lsa})
 reflects a wide range of authorial preferences, lexical, grammatical, and
 phonetic. While it can be clouded by external factors like genre and topic,
 it is a reliable and well-established approach. The poetic analysis (\S
 \ref{sec: poetic}) examined technical minutiae of prosody and metre that are
 standard considerations in classical philology; the innovation lies in their
 automatic extraction and their treatment with modern multivariate analysis
 and statistics. The consideration of some stylistic features related to rhyme
 is perhaps controversial, but my recent research \cite{nagy_rhyme_2022} has
 established that there is authorial signal in this domain. The poetic models
 performed very strongly, particularly in terms of differentiating authors,
 and it is in this domain that the separation between the Single and Double
 letters is most obvious.

 My interpretation of the results is quite simple: all twenty-one of the
 \emph{Heroides} are genuinely Ovidian, the \emph{ES} is in the correct
 position, and the Double \emph{Heroides} were written in exile. The analysis
 worked surprisingly well, considering the very small samples---this
 demonstrates the stylistic density of Latin poetry, and is cause for optimism
 regarding other questions of authorship. In this case, particularly because
 of the stylistic distinctiveness of the \emph{Heroides}---they form tight
 clusters---the results are quite clear. Other questions are less so. For
 example, the authenticity of \emph{Amores} 3.5 is still questioned by some; so
 far I can say that nothing in this analysis suggests that \emph{Am.} 3.5 is
 \emph{not} Ovidian (\emph{nihil obstat!}), but the \emph{Amores} show much
 more stylistic diversity than Ovid's other works, so some more focused
 analysis would be prudent. At the end of the day, computational stylometry is
 an exercise in statistics and probability: the data will not always provide a
 clear answer to every question, but in the case of the \emph{Heroides}, I
 suggest that it does.

\section{Future Work}

I conclude with some thoughts about work left undone. In the \emph{Heroides},
the authorship of two lengthy sections (16.39--144 and 21.145--248) have been
questioned.%
    \footnote{\citeA{kenney1979two} examines the passages in detail. He
    considers the literary arguments against to be insufficient.}
These sections are transmitted via a single witness which is late (1477), and
mechanically printed. Since the sections are long enough to be analysed as a
unit with the techniques used here, they could be isolated and examined
somehow. However, my view---which I have not formally tested---is that a
hundred or so lines of foreign material (a little less than a third and a half
of the poems respectively) would certainly have altered the stylistic
fingerprint of those poems at least to the point where they appeared unusual,
and so I accept them with only the slightest reservation. In other Ovidian
matters, above I mentioned \emph{Amores} 3.5, which seems worth investigating
further, since the debate has persisted into this century.%
    \footnote{E.g. \citeA{mckeown2013authenticity}, which also serves as a
    fine synthesis of the issues.}
The challenges will be the smaller size of the work (although 46 lines, based
on these results, now seems fairly reasonable) but more so the huge stylistic
variation encountered in the \emph{Amores}---the more things vary, the more
difficult it is to decide if something is atypical. Considering other authors,
there would no doubt be great interest in the long-debated elegies of Sulpicia
(which appear in the so-called \emph{Corpus Tibullianum})---the only surviving
(ancient) Latin verse written by a woman. I am not, however, at all
optimistic. The poems are tiny, and the undisputed Tibullan corpus is meagre
in comparison to Ovid.%
    \footnote{Tibullus is not the only alternative candidate that has been
    proposed, however a fair amount of the historical debate is little more
    than thinly-veiled sexism. Mathilda Skoie's \citeyear{skoie2002reading}
    monograph is a good place to begin further investigation.}
Finally, there remain many ways in which the analysis of non-lexical style
might be improved, such as measuring `complexity', register and
intertextuality, all of which are challenging problems in machine learning.
The methods used here are still evolving. Nonetheless, it is hoped that the
present work might contribute to furthering our understanding of this debated
and divisive Ovidian collection.


\footnotesize
\section*{Acknowledgements}

This work was supported by the Polish Academy of Sciences Grant 2020/39 / O /
HS2 / 02931.

\FloatBarrier
\normalsize
\clearpage
\bibliography{digilat}

\begin{thebibliography}{}

\bibitem [\protect \citeauthoryear {%
Barchiesi%
}{%
Barchiesi%
}{%
{\protect \APACyear {1996}}%
}]{%
barchiesi1996review}
\APACinsertmetastar {%
barchiesi1996review}%
\begin{APACrefauthors}%
Barchiesi, A.%
\end{APACrefauthors}%
\unskip\
\newblock
\APACrefYearMonthDay{1996}{}{}.
\newblock
{\BBOQ}\APACrefatitle {Review of {EJ} {K}enney (ed.), {O}vid {H}eroides
  {XVI--XXI} ({C}ambridge)} {Review of {EJ} {K}enney (ed.), {O}vid {H}eroides
  {XVI--XXI} ({C}ambridge)}.{\BBCQ}
\newblock
\APACjournalVolNumPages{Bryn Mawr Classical Review}{12}{}{}.
\PrintBackRefs{\CurrentBib}

\bibitem [\protect \citeauthoryear {%
Courtney%
}{%
Courtney%
}{%
{\protect \APACyear {1965}}%
}]{%
courtney_65}
\APACinsertmetastar {%
courtney_65}%
\begin{APACrefauthors}%
Courtney, E.%
\end{APACrefauthors}%
\unskip\
\newblock
\APACrefYearMonthDay{1965}{}{}.
\newblock
{\BBOQ}\APACrefatitle {{O}vidian and non-{O}vidian {H}eroides} {{O}vidian and
  non-{O}vidian {H}eroides}.{\BBCQ}
\newblock
\APACjournalVolNumPages{BICS}{}{12}{63--66}.
\newblock
\begin{APACrefURL} \url{http://www.jstor.org/stable/43646359} \end{APACrefURL}
\newblock
\begin{APACrefDOI} \doi{10.2307/43646359} \end{APACrefDOI}
\PrintBackRefs{\CurrentBib}

\bibitem [\protect \citeauthoryear {%
Courtney%
}{%
Courtney%
}{%
{\protect \APACyear {1997}}%
}]{%
courtney_97}
\APACinsertmetastar {%
courtney_97}%
\begin{APACrefauthors}%
Courtney, E.%
\end{APACrefauthors}%
\unskip\
\newblock
\APACrefYearMonthDay{1997}{}{}.
\newblock
{\BBOQ}\APACrefatitle {Echtheitskritik: {O}vidian and Non-{O}vidian {H}eroides
  Again} {Echtheitskritik: {O}vidian and non-{O}vidian {H}eroides
  again}.{\BBCQ}
\newblock
\APACjournalVolNumPages{CJ}{93}{2}{157--166}.
\newblock
\begin{APACrefURL} \url{http://www.jstor.org/stable/3298136} \end{APACrefURL}
\PrintBackRefs{\CurrentBib}

\bibitem [\protect \citeauthoryear {%
Davis%
}{%
Davis%
}{%
{\protect \APACyear {202x}}%
}]{%
davis_amores}
\APACinsertmetastar {%
davis_amores}%
\begin{APACrefauthors}%
Davis, P\BPBI J.%
\end{APACrefauthors}%
\unskip\
\newblock
\APACrefYear{202x}.
\newblock
\APACrefbtitle {Ovid {A}mores Book 3 Edited with Introduction, Translation, and
  Commentary} {Ovid {A}mores book 3 edited with introduction, translation, and
  commentary}.
\newblock
\APACaddressPublisher{}{OUP (forthcoming)}.
\PrintBackRefs{\CurrentBib}

\bibitem [\protect \citeauthoryear {%
Duhigg%
}{%
Duhigg%
}{%
{\protect \APACyear {1971}}%
}]{%
duhigg1971elegiac}
\APACinsertmetastar {%
duhigg1971elegiac}%
\begin{APACrefauthors}%
Duhigg, J.%
\end{APACrefauthors}%
\unskip\
\newblock
\APACrefYearMonthDay{1971}{}{}.
\newblock
{\BBOQ}\APACrefatitle {The elegiac metre of {C}atullus} {The elegiac metre of
  {C}atullus}.{\BBCQ}
\newblock
\APACjournalVolNumPages{Antichthon}{5}{}{57--67}.
\PrintBackRefs{\CurrentBib}

\bibitem [\protect \citeauthoryear {%
Dörrie%
}{%
Dörrie%
}{%
{\protect \APACyear {1971}}%
}]{%
dorrie_71}
\APACinsertmetastar {%
dorrie_71}%
\begin{APACrefauthors}%
Dörrie, H.%
\end{APACrefauthors}%
\ (\BED).
\unskip\
\newblock
\APACrefYear{1971}.
\newblock
\APACrefbtitle {Epistulae Heroidum} {Epistulae heroidum}.
\newblock
\APACaddressPublisher{}{de Grutyer}.
\PrintBackRefs{\CurrentBib}

\bibitem [\protect \citeauthoryear {%
Eder%
}{%
Eder%
}{%
{\protect \APACyear {2017}}%
}]{%
eder2017visualization}
\APACinsertmetastar {%
eder2017visualization}%
\begin{APACrefauthors}%
Eder, M.%
\end{APACrefauthors}%
\unskip\
\newblock
\APACrefYearMonthDay{2017}{}{}.
\newblock
{\BBOQ}\APACrefatitle {Visualization in stylometry: cluster analysis using
  networks} {Visualization in stylometry: cluster analysis using
  networks}.{\BBCQ}
\newblock
\APACjournalVolNumPages{Digital Scholarship in the Humanities}{32}{1}{50--64}.
\PrintBackRefs{\CurrentBib}

\bibitem [\protect \citeauthoryear {%
Fruchterman%
\ \BBA {} Reingold%
}{%
Fruchterman%
\ \BBA {} Reingold%
}{%
{\protect \APACyear {1991}}%
}]{%
fruchterman1991graph}
\APACinsertmetastar {%
fruchterman1991graph}%
\begin{APACrefauthors}%
Fruchterman, T\BPBI M.%
\BCBT {}\ \BBA {} Reingold, E\BPBI M.%
\end{APACrefauthors}%
\unskip\
\newblock
\APACrefYearMonthDay{1991}{}{}.
\newblock
{\BBOQ}\APACrefatitle {Graph drawing by force-directed placement} {Graph
  drawing by force-directed placement}.{\BBCQ}
\newblock
\APACjournalVolNumPages{Software: Practice and Experience}{21}{11}{1129--1164}.
\PrintBackRefs{\CurrentBib}

\bibitem [\protect \citeauthoryear {%
Heyworth%
}{%
Heyworth%
}{%
{\protect \APACyear {2016}}%
}]{%
heyworth_16}
\APACinsertmetastar {%
heyworth_16}%
\begin{APACrefauthors}%
Heyworth, S\BPBI J.%
\end{APACrefauthors}%
\unskip\
\newblock
\APACrefYearMonthDay{2016}{}{}.
\newblock
{\BBOQ}\APACrefatitle {Authenticity and Other Textual Problems in {H}eroides
  16} {Authenticity and other textual problems in {H}eroides 16}.{\BBCQ}
\newblock
\BIn{} R.~Hunter\ \BBA {} S\BPBI P.~Oakley\ (\BEDS), \APACrefbtitle {Latin
  Literature and its Transmission: Papers in Honour of {M}ichael {R}eeve}
  {Latin literature and its transmission: Papers in honour of {M}ichael
  {R}eeve}\ (\BPGS\ 142--170).
\newblock
\APACaddressPublisher{}{CUP}.
\PrintBackRefs{\CurrentBib}

\bibitem [\protect \citeauthoryear {%
Kenney%
}{%
Kenney%
}{%
{\protect \APACyear {1979}}%
}]{%
kenney1979two}
\APACinsertmetastar {%
kenney1979two}%
\begin{APACrefauthors}%
Kenney, E\BPBI J.%
\end{APACrefauthors}%
\unskip\
\newblock
\APACrefYearMonthDay{1979}{}{}.
\newblock
{\BBOQ}\APACrefatitle {Two disputed passages in the {H}eroides} {Two disputed
  passages in the {H}eroides}.{\BBCQ}
\newblock
\APACjournalVolNumPages{CQ}{29}{2}{394--431}.
\PrintBackRefs{\CurrentBib}

\bibitem [\protect \citeauthoryear {%
Kenney%
}{%
Kenney%
}{%
{\protect \APACyear {1996}}%
}]{%
kenney_96}
\APACinsertmetastar {%
kenney_96}%
\begin{APACrefauthors}%
Kenney, E\BPBI J.%
\end{APACrefauthors}%
\ (\BED).
\unskip\
\newblock
\APACrefYear{1996}.
\newblock
\APACrefbtitle {Ovid {H}eroides {XVI--XXI}} {Ovid {H}eroides {XVI--XXI}}.
\newblock
\APACaddressPublisher{}{CUP}.
\PrintBackRefs{\CurrentBib}

\bibitem [\protect \citeauthoryear {%
Kenney%
}{%
Kenney%
}{%
{\protect \APACyear {1999}}%
}]{%
kenney1999anomaly}
\APACinsertmetastar {%
kenney1999anomaly}%
\begin{APACrefauthors}%
Kenney, E\BPBI J.%
\end{APACrefauthors}%
\unskip\
\newblock
\APACrefYearMonthDay{1999}{}{}.
\newblock
{\BBOQ}\APACrefatitle {Anomaly, Innovation and Genre in {O}vid, {H}eroides
  16--21} {Anomaly, innovation and genre in {O}vid, {H}eroides 16--21}.{\BBCQ}
\newblock
\APACjournalVolNumPages{Aspects of the Language of Latin Poetry (Proceedings of
  the British Academy)}{93}{}{399--414}.
\PrintBackRefs{\CurrentBib}

\bibitem [\protect \citeauthoryear {%
Koppel%
, Schler%
, Argamon%
\BCBL {}\ \BBA {} Winter%
}{%
Koppel%
\ \protect \BOthers {.}}{%
{\protect \APACyear {2012}}%
}]{%
koppel_fundamental_2012}
\APACinsertmetastar {%
koppel_fundamental_2012}%
\begin{APACrefauthors}%
Koppel, M.%
, Schler, J.%
, Argamon, S.%
\BCBL {}\ \BBA {} Winter, Y.%
\end{APACrefauthors}%
\unskip\
\newblock
\APACrefYearMonthDay{2012}{{\APACmonth{05}}}{}.
\newblock
{\BBOQ}\APACrefatitle {The “{Fundamental} {Problem}” of {Authorship}
  {Attribution}} {The “{Fundamental} {Problem}” of {Authorship}
  {Attribution}}.{\BBCQ}
\newblock
\APACjournalVolNumPages{English Studies}{93}{3}{284--291}.
\newblock
\begin{APACrefURL}
  [{2019-10-16}]\url{http://www.tandfonline.com/doi/abs/10.1080/0013838X.2012.668794}
  \end{APACrefURL}
\newblock
\begin{APACrefDOI} \doi{10.1080/0013838X.2012.668794} \end{APACrefDOI}
\PrintBackRefs{\CurrentBib}

\bibitem [\protect \citeauthoryear {%
McInnes%
, Healy%
\BCBL {}\ \BBA {} Melville%
}{%
McInnes%
\ \protect \BOthers {.}}{%
{\protect \APACyear {2018}}%
}]{%
mcinnes_umap_2018}
\APACinsertmetastar {%
mcinnes_umap_2018}%
\begin{APACrefauthors}%
McInnes, L.%
, Healy, J.%
\BCBL {}\ \BBA {} Melville, J.%
\end{APACrefauthors}%
\unskip\
\newblock
\APACrefYearMonthDay{2018}{{\APACmonth{12}}}{}.
\newblock
{\BBOQ}\APACrefatitle {{UMAP}: {Uniform} {Manifold} {Approximation} and
  {Projection} for {Dimension} {Reduction}} {{UMAP}: {Uniform} {Manifold}
  {Approximation} and {Projection} for {Dimension} {Reduction}}.{\BBCQ}
\newblock
\APACjournalVolNumPages{arXiv:1802.03426 [cs, stat]}{}{}{}.
\newblock
\begin{APACrefURL} [{2020-06-18}]\url{http://arxiv.org/abs/1802.03426}
  \end{APACrefURL}
\PrintBackRefs{\CurrentBib}

\bibitem [\protect \citeauthoryear {%
McKeown%
}{%
McKeown%
}{%
{\protect \APACyear {1998}}%
}]{%
mckeown_98}
\APACinsertmetastar {%
mckeown_98}%
\begin{APACrefauthors}%
McKeown, J\BPBI C.%
\end{APACrefauthors}%
\unskip\
\newblock
\APACrefYear{1998}.
\newblock
\APACrefbtitle {Ovid {A}mores: {T}ext, {P}rolegomena and {C}ommentary} {Ovid
  {A}mores: {T}ext, {P}rolegomena and {C}ommentary}\ (\BVOL~III).
\newblock
\APACaddressPublisher{}{Francis Cairns}.
\PrintBackRefs{\CurrentBib}

\bibitem [\protect \citeauthoryear {%
McKeown%
}{%
McKeown%
}{%
{\protect \APACyear {2013}}%
}]{%
mckeown2013authenticity}
\APACinsertmetastar {%
mckeown2013authenticity}%
\begin{APACrefauthors}%
McKeown, J\BPBI C.%
\end{APACrefauthors}%
\unskip\
\newblock
\APACrefYearMonthDay{2013}{}{}.
\newblock
{\BBOQ}\APACrefatitle {The authenticity of {A}mores 3.5} {The authenticity of
  {A}mores 3.5}.{\BBCQ}
\newblock
\BIn{} \APACrefbtitle {Vertis in usum} {Vertis in usum}\ (\BPGS\ 114--128).
\newblock
\APACaddressPublisher{}{BG Teubner}.
\PrintBackRefs{\CurrentBib}

\bibitem [\protect \citeauthoryear {%
Morgan%
}{%
Morgan%
}{%
{\protect \APACyear {1977}}%
}]{%
morgan1977ovid}
\APACinsertmetastar {%
morgan1977ovid}%
\begin{APACrefauthors}%
Morgan, K.%
\end{APACrefauthors}%
\unskip\
\newblock
\APACrefYear{1977}.
\newblock
\APACrefbtitle {Ovid's Art of Imitation: {P}ropertius in the {A}mores} {Ovid's
  art of imitation: {P}ropertius in the {A}mores}\ (\BVOL~47).
\newblock
\APACaddressPublisher{}{Brill Mnemosyne Supplements}.
\PrintBackRefs{\CurrentBib}

\bibitem [\protect \citeauthoryear {%
Nagy%
}{%
Nagy%
}{%
{\protect \APACyear {2021}}%
}]{%
nagy2021metre}
\APACinsertmetastar {%
nagy2021metre}%
\begin{APACrefauthors}%
Nagy, B.%
\end{APACrefauthors}%
\unskip\
\newblock
\APACrefYearMonthDay{2021}{}{}.
\newblock
{\BBOQ}\APACrefatitle {Metre as a stylometric feature in {L}atin hexameter
  poetry} {Metre as a stylometric feature in {L}atin hexameter poetry}.{\BBCQ}
\newblock
\APACjournalVolNumPages{Digital Scholarship in the
  Humanities}{36}{4}{999--1012}.
\newblock
\begin{APACrefDOI} \doi{10.1093/llc/fqaa043} \end{APACrefDOI}
\PrintBackRefs{\CurrentBib}

\bibitem [\protect \citeauthoryear {%
Nagy%
}{%
Nagy%
}{%
{\protect \APACyear {2022}}%
{\protect \APACexlab {{\protect \BCnt {1}}}}}]{%
nagy_heroides_2022}
\APACinsertmetastar {%
nagy_heroides_2022}%
\begin{APACrefauthors}%
Nagy, B.%
\end{APACrefauthors}%
\unskip\
\newblock
\APACrefYearMonthDay{2022{\protect \BCnt {1}}}{}{}.
\newblock
{\BBOQ}\APACrefatitle {Preprint: Some Stylometric Remarks On {O}vid’s
  \emph{{H}eroides} And The \emph{{E}pistula {S}apphus}} {Preprint: Some
  stylometric remarks on {O}vid’s \emph{{H}eroides} and the \emph{{E}pistula
  {S}apphus}}.{\BBCQ}
\newblock
\APAChowpublished {https://github.com/bnagy/heroides-paper}.
\newblock
\begin{APACrefDOI} \doi{10.5281/zenodo.6099889} \end{APACrefDOI}
\PrintBackRefs{\CurrentBib}

\bibitem [\protect \citeauthoryear {%
Nagy%
}{%
Nagy%
}{%
{\protect \APACyear {2022}}%
{\protect \APACexlab {{\protect \BCnt {2}}}}}]{%
nagy_rhyme_2022}
\APACinsertmetastar {%
nagy_rhyme_2022}%
\begin{APACrefauthors}%
Nagy, B.%
\end{APACrefauthors}%
\unskip\
\newblock
\APACrefYearMonthDay{2022{\protect \BCnt {2}}}{}{}.
\newblock
{\BBOQ}\APACrefatitle {{R}hyme in classical {L}atin poetry: stylistic or
  stochastic?} {{R}hyme in classical {L}atin poetry: stylistic or
  stochastic?}{\BBCQ}
\newblock
\APACjournalVolNumPages{Digital Scholarship in the Humanities}{}{}{(in press)}.
\newblock
\begin{APACrefDOI} \doi{10.1093/llc/fqab105} \end{APACrefDOI}
\PrintBackRefs{\CurrentBib}

\bibitem [\protect \citeauthoryear {%
Pedregosa%
\ \protect \BOthers {.}}{%
Pedregosa%
\ \protect \BOthers {.}}{%
{\protect \APACyear {2011}}%
}]{%
scikit-learn}
\APACinsertmetastar {%
scikit-learn}%
\begin{APACrefauthors}%
Pedregosa, F.%
, Varoquaux, G.%
, Gramfort, A.%
, Michel, V.%
, Thirion, B.%
, Grisel, O.%
\BDBL {}Duchesnay, E.%
\end{APACrefauthors}%
\unskip\
\newblock
\APACrefYearMonthDay{2011}{}{}.
\newblock
{\BBOQ}\APACrefatitle {Scikit-learn: Machine Learning in {P}ython}
  {Scikit-learn: Machine learning in {P}ython}.{\BBCQ}
\newblock
\APACjournalVolNumPages{Journal of Machine Learning
  Research}{12}{}{2825--2830}.
\PrintBackRefs{\CurrentBib}

\bibitem [\protect \citeauthoryear {%
Reeve%
}{%
Reeve%
}{%
{\protect \APACyear {1973}}%
}]{%
reeve_heroides}
\APACinsertmetastar {%
reeve_heroides}%
\begin{APACrefauthors}%
Reeve, M\BPBI D.%
\end{APACrefauthors}%
\unskip\
\newblock
\APACrefYearMonthDay{1973}{}{}.
\newblock
{\BBOQ}\APACrefatitle {Notes on {O}vid's {H}eroides} {Notes on {O}vid's
  {H}eroides}.{\BBCQ}
\newblock
\APACjournalVolNumPages{CQ}{23}{2}{324--338}.
\newblock
\begin{APACrefURL} \url{http://www.jstor.org/stable/638190} \end{APACrefURL}
\newblock
\begin{APACrefDOI} \doi{10.2307/638190} \end{APACrefDOI}
\PrintBackRefs{\CurrentBib}

\bibitem [\protect \citeauthoryear {%
Skoie%
}{%
Skoie%
}{%
{\protect \APACyear {2002}}%
}]{%
skoie2002reading}
\APACinsertmetastar {%
skoie2002reading}%
\begin{APACrefauthors}%
Skoie, M.%
\end{APACrefauthors}%
\unskip\
\newblock
\APACrefYear{2002}.
\newblock
\APACrefbtitle {Reading {S}ulpicia: Commentaries, 1475-1990} {Reading
  {S}ulpicia: Commentaries, 1475-1990}.
\newblock
\APACaddressPublisher{}{OUP}.
\PrintBackRefs{\CurrentBib}

\bibitem [\protect \citeauthoryear {%
Tarrant%
}{%
Tarrant%
}{%
{\protect \APACyear {1981}}%
}]{%
tarrant_81}
\APACinsertmetastar {%
tarrant_81}%
\begin{APACrefauthors}%
Tarrant, R\BPBI J.%
\end{APACrefauthors}%
\unskip\
\newblock
\APACrefYearMonthDay{1981}{}{}.
\newblock
{\BBOQ}\APACrefatitle {The Authenticity of the Letter of {Sappho} to {Phaon}
  ({Heroides XV})} {The authenticity of the letter of {Sappho} to {Phaon}
  ({Heroides XV})}.{\BBCQ}
\newblock
\APACjournalVolNumPages{Harvard Studies in Classical
  Philology}{85}{}{133--153}.
\newblock
\begin{APACrefURL} \url{http://www.jstor.org/stable/311169} \end{APACrefURL}
\newblock
\begin{APACrefDOI} \doi{10.2307/311169} \end{APACrefDOI}
\PrintBackRefs{\CurrentBib}

\bibitem [\protect \citeauthoryear {%
Tarrant%
}{%
Tarrant%
}{%
{\protect \APACyear {1983}}%
}]{%
tarrant_trans_ep}
\APACinsertmetastar {%
tarrant_trans_ep}%
\begin{APACrefauthors}%
Tarrant, R\BPBI J.%
\end{APACrefauthors}%
\unskip\
\newblock
\APACrefYearMonthDay{1983}{}{}.
\newblock
{\BBOQ}\APACrefatitle {Ovid, {Heroides}} {Ovid, {Heroides}}.{\BBCQ}
\newblock
\BIn{} L\BPBI D.~Reynolds\ (\BED), \APACrefbtitle {Texts and {Transmission}}
  {Texts and {Transmission}}\ (\BPGS\ 268--73).
\newblock
\APACaddressPublisher{}{OUP}.
\PrintBackRefs{\CurrentBib}

\bibitem [\protect \citeauthoryear {%
Thorsen%
}{%
Thorsen%
}{%
{\protect \APACyear {2014}}%
}]{%
thorsen_ovearly}
\APACinsertmetastar {%
thorsen_ovearly}%
\begin{APACrefauthors}%
Thorsen, T.%
\end{APACrefauthors}%
\unskip\
\newblock
\APACrefYear{2014}.
\newblock
\APACrefbtitle {{O}vid's Early Poetry: from his {S}ingle {H}eroides to his
  {R}emedia {A}moris} {{O}vid's early poetry: from his {S}ingle {H}eroides to
  his {R}emedia {A}moris}.
\newblock
\APACaddressPublisher{}{CUP}.
\newblock
\begin{APACrefDOI} \doi{10.1017/CBO9781139628952} \end{APACrefDOI}
\PrintBackRefs{\CurrentBib}

\bibitem [\protect \citeauthoryear {%
Van~der Maaten%
\ \BBA {} Hinton%
}{%
Van~der Maaten%
\ \BBA {} Hinton%
}{%
{\protect \APACyear {2008}}%
}]{%
van2008visualizing}
\APACinsertmetastar {%
van2008visualizing}%
\begin{APACrefauthors}%
Van~der Maaten, L.%
\BCBT {}\ \BBA {} Hinton, G.%
\end{APACrefauthors}%
\unskip\
\newblock
\APACrefYearMonthDay{2008}{}{}.
\newblock
{\BBOQ}\APACrefatitle {Visualizing data using t-{SNE}.} {Visualizing data using
  t-{SNE}.}{\BBCQ}
\newblock
\APACjournalVolNumPages{Journal of Machine Learning Research}{9}{11}{}.
\PrintBackRefs{\CurrentBib}

\bibitem [\protect \citeauthoryear {%
West%
}{%
West%
}{%
{\protect \APACyear {1957}}%
}]{%
west1957metre}
\APACinsertmetastar {%
west1957metre}%
\begin{APACrefauthors}%
West, D\BPBI A.%
\end{APACrefauthors}%
\unskip\
\newblock
\APACrefYearMonthDay{1957}{}{}.
\newblock
{\BBOQ}\APACrefatitle {The Metre Of {C}atullus' Elegiacs} {The metre of
  {C}atullus' elegiacs}.{\BBCQ}
\newblock
\APACjournalVolNumPages{CQ}{7}{1-2}{98--102}.
\PrintBackRefs{\CurrentBib}

\bibitem [\protect \citeauthoryear {%
Wickham%
}{%
Wickham%
}{%
{\protect \APACyear {2016}}%
}]{%
ggplot}
\APACinsertmetastar {%
ggplot}%
\begin{APACrefauthors}%
Wickham, H.%
\end{APACrefauthors}%
\unskip\
\newblock
\APACrefYear{2016}.
\newblock
\APACrefbtitle {ggplot2: Elegant Graphics for Data Analysis} {ggplot2: Elegant
  graphics for data analysis}.
\newblock
\APACaddressPublisher{}{Springer-Verlag New York}.
\newblock
\begin{APACrefURL} \url{https://ggplot2.tidyverse.org} \end{APACrefURL}
\PrintBackRefs{\CurrentBib}

\bibitem [\protect \citeauthoryear {%
Wilkinson%
}{%
Wilkinson%
}{%
{\protect \APACyear {1955}}%
}]{%
wilkinson_74}
\APACinsertmetastar {%
wilkinson_74}%
\begin{APACrefauthors}%
Wilkinson, L\BPBI P.%
\end{APACrefauthors}%
\unskip\
\newblock
\APACrefYear{1955}.
\newblock
\APACrefbtitle {Ovid {R}ecalled} {Ovid {R}ecalled}.
\newblock
\APACaddressPublisher{}{Cedric Chivers}.
\newblock
\APACrefnote{repr. 1974}
\PrintBackRefs{\CurrentBib}

\bibitem [\protect \citeauthoryear {%
Zimek%
, Schubert%
\BCBL {}\ \BBA {} Kriegel%
}{%
Zimek%
\ \protect \BOthers {.}}{%
{\protect \APACyear {2012}}%
}]{%
zimek_etal}
\APACinsertmetastar {%
zimek_etal}%
\begin{APACrefauthors}%
Zimek, A.%
, Schubert, E.%
\BCBL {}\ \BBA {} Kriegel, H\BHBI P.%
\end{APACrefauthors}%
\unskip\
\newblock
\APACrefYearMonthDay{2012}{}{}.
\newblock
{\BBOQ}\APACrefatitle {A survey on unsupervised outlier detection in
  high-dimensional numerical data} {A survey on unsupervised outlier detection
  in high-dimensional numerical data}.{\BBCQ}
\newblock
\APACjournalVolNumPages{Statistical Analysis and Data Mining}{5}{5}{363--387}.
\newblock
\begin{APACrefDOI} \doi{https://doi.org/10.1002/sam.11161} \end{APACrefDOI}
\PrintBackRefs{\CurrentBib}

\end{thebibliography}


\end{document}